\documentclass{article}

\usepackage{amssymb}
\usepackage{amsmath}
\usepackage{graphicx}
\usepackage{float}
\usepackage{booktabs}
\usepackage{multirow}
\usepackage[table]{xcolor}
\usepackage{array}
\usepackage{wrapfig}
\usepackage{xspace}
\usepackage[normalem]{ulem}
\usepackage{fontawesome5} 
\usepackage{tcolorbox}
\usepackage{hyperref}
\usepackage{xcolor}
\usepackage{caption}
\usepackage{algorithm}
\usepackage{algpseudocode}

\usepackage[final]{corl_2026} %

\newcommand{\name}{PartialBiGrasp\xspace}

\title{\name: Inferring Hidden Local Geometry for Bimanual Grasping from Partial Views}

\author{
  Ayush Kaura$^{*}$, Vignesh Vembar$^{*}$, Md Faizal Karim,
  Keshab Patra, K Madhava Krishna \\
  \normalfont Robotics Research Center, IIIT Hyderabad, \normalfont $^{*}$Equal contribution.
}

\usepackage{xcolor}

\definecolor{codebg}{RGB}{248,248,248}
\definecolor{codecolor}{RGB}{140,142,144}

\begin{document}
\maketitle

\definecolor{red}{RGB}{255, 77, 109}
\definecolor{green}{RGB}{70, 171, 43}

\vspace{-20pt}
\begin{figure*}[h]
    \centering
    \includegraphics[width=\textwidth]{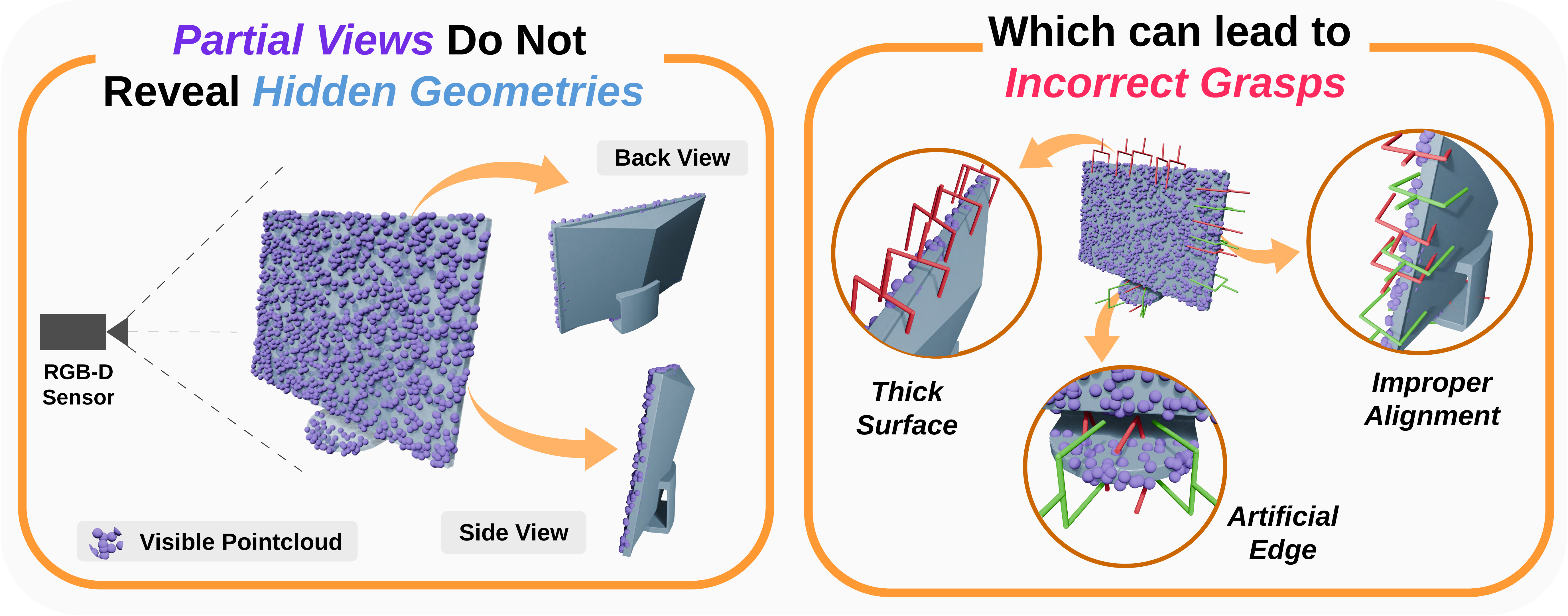}
    \caption{\small Partial observations hide grasp-relevant geometry and introduce deceptive grasp cues, often leading to unstable grasps, as denoted by the \textcolor{red}{red} grasps. PartialBiGrasp overcomes these challenges, shown as the \textcolor{green}{green} grasps, by reasoning about hidden local geometry to generate physically feasible dual-arm grasps.}
    \label{fig:teaser}
    \vspace{-2mm}
\end{figure*}

\begin{abstract}
    Dual-arm robotic grasping is essential for manipulating large, heavy, and geometrically complex objects that cannot be reliably handled using a single manipulator. These large objects often contain only sparse graspable regions determined by local geometric properties such as thickness, edge structure, and gripper clearance. 
    Prior bimanual grasping methods assume access to a full point cloud of the object which inherently contains this geometric information, but may not be accessible in real scenarios.
    This work proposes \name, a dual-arm grasp generation framework that operates directly on partial point cloud observations.
    Our model learns geometric features implicitly through convolutional occupancy networks, enabling local reasoning about graspability, collision-free contact regions, and object thickness. We leverage this understanding to generate force-closure compliant grasp pairs, which are further refined using a sampling-based optimization to correct for ambiguity caused by incomplete geometry. 
    We evaluate our approach using analytical force-closure metrics, large-scale simulation experiments, and real-world robot evaluations on noisy partial point clouds of novel objects, demonstrating robust and physically stable dual-arm grasp generation. Project Page: \url{https://partialbigrasp.github.io}
    
\end{abstract}


\section{Introduction}
\label{sec:introduction}

Everyday tasks such as carrying boxes, lifting trays, and moving chairs naturally require coordinated interaction between two robotic arms to support, stabilize, and manipulate them \cite{dual_arm_manipulation, robot_learning_survey}. By distributing contacts between both manipulators, bimanual grasping improves object grasp stability and load balancing, enabling robust manipulation of these bulky objects that would otherwise induce unstable interactions for a single arm~\cite{robotic_grasping_classical}. Despite its importance, relatively few works have explored learning-based bimanual grasp generation, particularly under partial observation settings.

Recent years have witnessed substantial progress in single-arm robotic grasp synthesis driven by learning-based methods \cite{survey2023deepgraspsynthesis, contact_graspnet, 6dof_graspnet, giga, anygrasp, se3diff, graspgen}. However, translating these advances to bimanual grasping remains challenging. Firstly, a valid dual-arm grasp is not simply a combination of two independent grasps, instead, both grasps must jointly satisfy physical force-closure constraints~\cite{dg16m}. Secondly, unlike small tabletop objects, large objects are often graspable only at sparse local regions such as thin edges, handles, or corners, where grasp feasibility depends critically on fine-grained local geometry including object thickness and gripper clearance. Finally, these challenges become significantly harder under realistic single-view partial observations, where stable grasp generation requires reasoning over hidden local geometry together with global geometric understanding to accurately predict force-balanced grasp pairs.

Existing approaches such as DAGDiff \cite{dagdiff} employ a diffusion model to learn the multimodal distribution of bimanual grasps, while BiGraspFormer \cite{bigraspformer} directly predicts grasp pairs using a transformer-based architecture. However, these methods require complete object geometry from dense or fused point cloud observations, and fail under single-view partial observations due to their limited understanding of hidden local geometry. They are further hindered by the fact that incomplete views can introduce artificial graspable surfaces that appear geometrically valid despite being physically infeasible, as shown in Figure \ref{fig:teaser}.

To address these challenges, we develop \textbf{PartialBiGrasp}, a framework for generating stable dual-arm grasps directly from a single RGB-D observation. Rather than relying on complete object geometry or explicit shape reconstruction, our method reasons about hidden local geometry to predict force-balanced and collision-free grasp pairs for large objects with sparse graspable regions.

Our approach leverages Convolutional OccNets \cite{conv_occ_net} to learn occupancy-conditioned geometric representations that enable reasoning beyond visible point cloud observations. Unlike previous methods that use partial point clouds, our framework learns to reason
about hidden local geometry beyond visible surfaces, which is critical for understanding grasp-relevant properties such as local thickness and surface continuation under occlusion. In contrast to explicit point cloud completion or single-view reconstruction approaches \cite{yu2021pointr, sam3d2025, recgen}, which can introduce inaccurate local geometry around thin structures and contact regions while additionally requiring stagewise reconstruction-and-grasping pipelines, our method directly learns grasp-oriented geometric reasoning in a unified framework. We first learn global geometric features to identify feasible grasp regions and predict compatible dual-arm grasp pairs satisfying force-closure constraints. To further improve grasp alignment around contact regions, we additionally learn grasp-aligned local occupancy features that encode fine-grained geometric properties enabling collision-free grasp refinement under incomplete observations. To summarize the contributions:

\begin{enumerate} %
    \item We propose \textbf{\name}, the first framework to the best of our knowledge, for bimanual grasp generation on single-view partial point clouds. It decomposes the task into single-grasp generation, force-closure based grasp pairing, and local geometry conditioned refinement, enabling efficient generation of stable dual-arm grasps under partial observations.
    \item We introduce an occupancy-conditioned geometric reasoning pipeline that leverages convolutional occupancy networks and grasp-aligned local features to infer graspability, local object geometry, collision-free contact regions, and gripper feasibility, together with a sampling-based refinement procedure to improve grasp quality.
    \item We evaluate our approach through analytical force-closure validation, large-scale simulation experiments, and real-world robot deployment, demonstrating robust dual-arm grasp generation across diverse large-object scenarios.
\end{enumerate}

\section{Related Works}
\label{sec:related-works}

\textbf{Grasping under Partial Observations.} Grasp synthesis from visual observations can be broadly classified into three categories. Direct predictors \cite{gpd, contact_graspnet, anygrasp, 6dof_graspnet} take noisy partial RGB-D data as input and output feasible grasp poses, making them attractive for real-world deployment. However, these methods are primarily designed for small tabletop objects and rely on dense grasp supervision that exists at small object scales. A second line of work addresses partial observations through geometry reconstruction, where methods such as \cite{giga, 3dgs_grasp, scene_grasp, zero_grasp} first infer missing object geometry before grasp generation. While this enables reasoning over occluded regions, successful grasping depends on accurately reconstructing hidden geometry, which becomes increasingly difficult for large objects due to complex structures. More recent generative methods \cite{se3diff, graspgen, cgdf} better capture the multimodal nature of grasping and can handle more complex object geometries, but typically rely on complete object geometry supervision. In contrast, our method operates directly on single-view observations, leveraging an implicit local geometry representation to generate physically feasible grasp pairs without reconstructing the complete object.

\textbf{Dual-Arm Grasp Generation.} Dual-arm grasping requires two grasps that are (i) individually valid and (ii) jointly stable under force-closure constraints \cite{force-closure}. Sampling-based approaches \cite{dg16m, da2} densely sample single-arm grasps on object meshes, pair them, and filter feasible combinations using stability metrics. Although effective, the requirement of precise meshes makes it impractical for real-world deployment. To overcome this, recent generative approaches have adopted learning-based generation strategies. \cite{cgdf} uses diffusion models to generate single-arm grasps and pairs them heuristically, while  \cite{dagdiff} directly generates coordinated grasp pairs using guided diffusion. More recently, \cite{bigraspformer} %
employs a transformer to directly predict grasp pairs from learnable queries \cite{detr}. Recent works have also explored VLM-based semantic reasoning for pairing dual-arm grasps and affordance prediction \cite{unidiff, taskawarebimanualaffordanceprediction}. Our framework directly generates stable grasp pairs using a feed-forward style network, without requiring meshes, heuristic pairing strategies, or access to VLMs.

\section{Methods}
\label{sec:methods}

Given a single RGB-D observation of an object $\mathcal{I} \in \mathbb{R}^{H\times W \times 4}$ along with its segmentation mask, our aim is to generate \textit{M} pairs of bimanual parallel jaw gripper poses $\left\{(G_{i1}, G_{i2}) \in SE(3) \times SE(3) \right\}_{i=1}^{M}$ such that the resulting grasp pairs are force-closure stable and collision-free. Following \cite{contact_graspnet}, we assume that at least one contact point of each grasp is visible in the partial observation.

\begin{figure}[!t]
    \centering
    \includegraphics[width=1\linewidth]{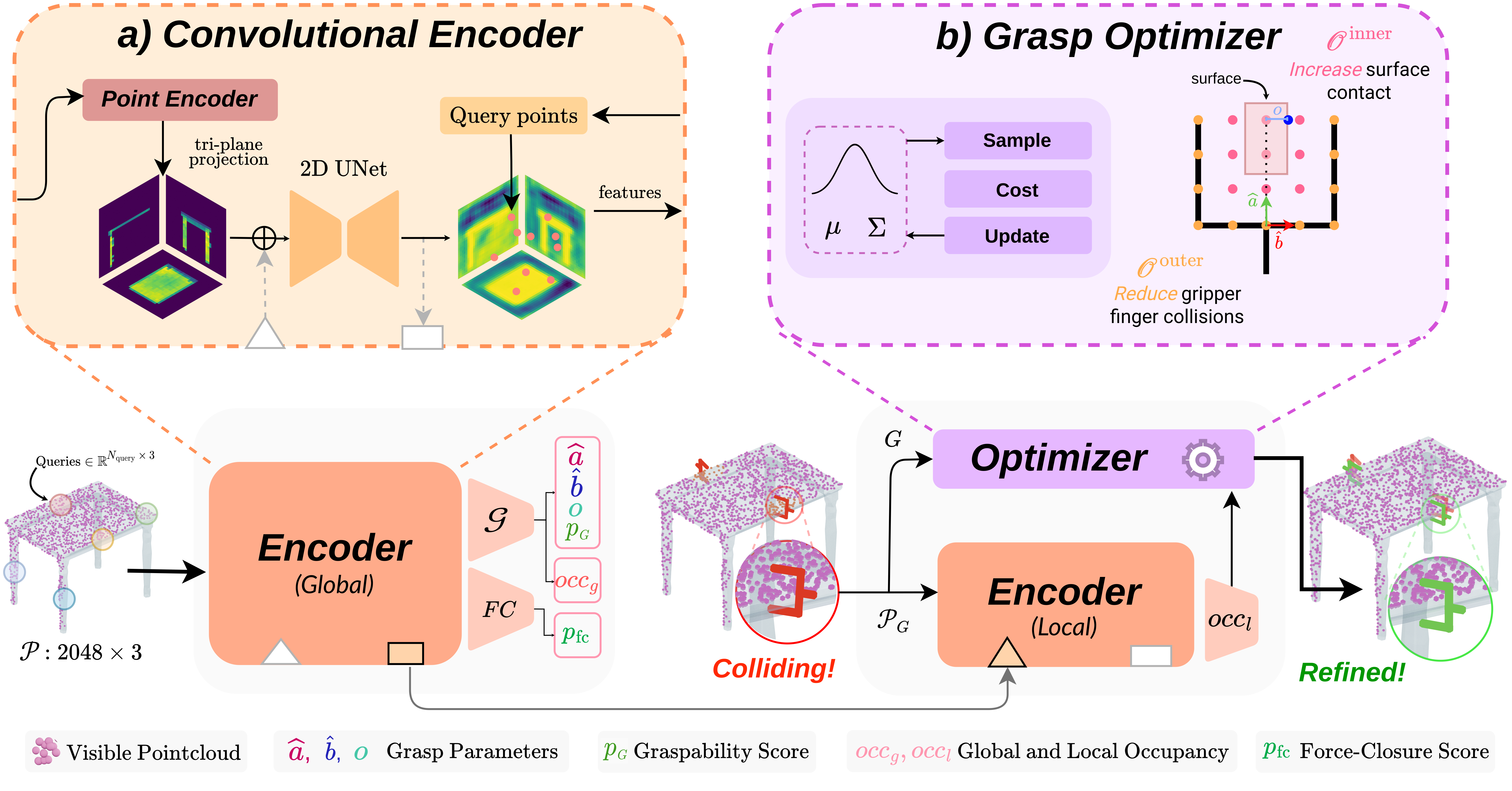}
    \caption{\small Given a partial point cloud $\mathcal{P}$, the Global Encoder produces a continuous object representation from which the Grasp Generation Module $\mathcal{G}$ decodes single-arm grasps. High-graspability candidates are paired and scored by the Force Closure Module $FC$. Local geometry is captured by encoding gripper-frame point cloud crops $\mathcal{P}_G$ with the Local Encoder. Finally, a sampling-based optimizer uses local occupancy predictions $occ_l$ to reduce collisions and enforce stable contact, yielding force-closure-stable grasps that are aware of hidden local geometry.}
    \label{fig:pipeline}
\end{figure}

Using $\mathcal{I}$ and the object segmentation mask, we first project the object into a point cloud $\mathcal{P} \in \mathbb{R}^{n \times 3}$ and learn an implicit geometry representation that captures global object structure. This representation is used to generate candidate single-arm grasps and identify force-closure stable bimanual grasp pairs. Finally, we refine the predicted grasps using local occupancy-guided optimization to improve contact quality and reduce collisions.
The central idea of our approach is to reason about hidden local geometry from partial observations, enabling robust dual-arm grasp generation as shown in Figure \ref{fig:pipeline}. We describe each of these components in the following subsections.

\subsection{Implicit Geometry Representation}

The key challenge in single-view grasping is that grasp feasibility often depends on geometric properties that are not directly observed. For example, determining whether a visible edge can accommodate a parallel-jaw gripper requires reasoning about hidden quantities such as local thickness, surface continuation, and gripper clearance. To infer such grasp-relevant geometry from partial observations, we learn an implicit geometry representation that captures both global object structure and fine-grained local shape information.
As depicted in Figure \ref{fig:pipeline} (a), we extract rotation-equivariant point features of $\mathcal{P}$ using a point-based encoder~\cite{vnn}. 
To enable feature extraction at arbitrary 3D locations beyond the discrete object point cloud, the point features are projected onto a triplane representation and processed using a shared 2D U-Net~\cite{unet} to aggregate multi-scale spatial context \cite{conv_occ_net}.
The resulting representation defines a continuous feature field that can be queried at arbitrary 3D locations via interpolation.

We employ two complementary encoders built on this representation, each designed to capture geometry at a different spatial scale. The \textbf{Global Geometry Encoder} operates on the entire observed point cloud $\mathcal{P}$ and learns object-level geometric structure, enabling reasoning about overall shape, graspable regions, and the spatial relationships between potential grasp locations. In contrast, the \textbf{Local Geometry Encoder} focuses on geometry in the vicinity of a candidate grasp. Operating within a grasp-centered coordinate frame, it learns local occupancy features that encode cues relevant to grasp feasibility such as local thickness, surface continuation, and gripper clearance. Moreover, to preserve awareness of the overall object structure, the Local Encoder is conditioned on features from the Global Encoder via a \textit{residual} connection. Both encoders are trained by sampling 3D query points either globally around the object or locally around candidate grasp regions, and supervising their occupancy using a BCE loss. Together, the two encoders provide both the global context required for dual-arm grasp synthesis and the local geometric detail necessary for contact reasoning and grasp refinement.

\subsection{Dual-Arm Grasp Synthesis}

Using the implicit geometry representation described above, we generate dual-arm grasps in two stages. First, we predict feasible single-arm grasp candidates from the observed object geometry and then evaluate candidate grasp pairs using a learned force-closure critic.

\vspace{-2mm}
\paragraph{Single-Grasp Candidate Generation.}
For each candidate contact location in $\mathcal{P}$, a fixed set of canonical query points $\mathcal{P}_{\text{query}} \in \mathbb{R}^{N_{\text{query}} \times 3}$, defined as a spherical neighborhood around the contact point, is translated to the corresponding surface region and evaluated using the Global Encoder. The resulting features capture both the coarse surface geometry and object-level context surrounding the grasp location. 

The queried features are projected to a latent grasp descriptor via $\mathcal{G}$ and processed by multiple parallel prediction heads. (i) The \textbf{Graspability Head} predicts whether the queried contact location corresponds to a feasible grasp and is trained using a BCE loss over sparsely sampled positive and negative contact points from the dataset (as shown in Appendix \ref{subsec:Dataset}), enabling it to learn graspable object regions from sparse supervision.
(ii) The \textbf{Pose Regression Heads} predict the grasp parameters, namely the approach vector ($\boldsymbol{\hat{a}}$), baseline vector ($\boldsymbol{\hat{b}}$), and gripper offset (${o}$). They are supervised using $L_1$ losses on the gripper control points (depicted in orange in Figure \ref{fig:pipeline} (b)) and the gripper offset. Only positive grasp samples contribute to the pose regression heads' weight updates.

\vspace{-2mm}
\paragraph{Dual-Grasp Critic.}
Not all individually feasible grasps form a stable bimanual grasp when combined. To reason about cooperative stability, we construct candidate grasp pairs from the generated single-arm grasps and evaluate them using a learned force-closure critic.

For each grasp candidate, the canonical query points are translated using the predicted grasp pose and re-evaluated through the Global Encoder, and the features are projected to a latent grasp descriptor via $FC$. The representations corresponding to a candidate pair $(G_1, G_2)$ are concatenated and passed to a critic network that predicts whether the pair forms a force-closure stable dual-arm grasp. To encourage order invariance, the left and right grasp representations are randomly swapped during training with probability $0.5$. The critic is trained as a binary classifier using Force Closure positive and negative grasp pairs from DG16M \cite{dg16m}.

\subsection{Local Occupancy-Guided Grasp Refinement}

\begin{wrapfigure}{r}{0.45\columnwidth}
    \centering
    \vspace{-10pt}
    \includegraphics[width=0.45\columnwidth]{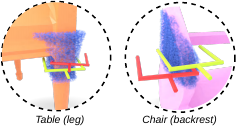}
    \caption{\small Local Occupancy Guided Refinement implicitly understands unknown properties such as local thickness and surface continuation under occlusion, and successfully updates grasp poses in local regions, maintaining grasp pair configuration while reducing collisions. Red signifies the initial pose, and Green signifies the refined pose.}
    \vspace{-20pt}
    \label{fig:refinement}
\end{wrapfigure}

Although the grasp synthesis stage generates force-closure stable dual-arm grasp pairs, the predicted poses may still exhibit local geometric inconsistencies arising due to the nature of partial observations as seen in Figure \ref{fig:teaser}. To address these errors, we perform a lightweight occupancy-guided refinement process using the Local Geometry Encoder, where grasp poses are iteratively improved through a sampling based optimization procedure.

Given the predicted dual-arm grasp pairs, we extract the unique set of single-arm grasps and refine each independently. Refinement is formulated as an optimization problem in the Lie algebra $\mathfrak{se}(3)$, where small perturbations are sampled around the current grasp pose and evaluated using the local occupancy field as seen in Figure \ref{fig:refinement}. Since the optimization is restricted to a small neighborhood of the predicted grasp, refinement primarily corrects local geometric inconsistencies such as minor collisions and suboptimal contact placement without moving grasps to different contact regions.

For each candidate grasp pose, we first extract a local partial point cloud around the gripper and encode it using the Local Geometry Encoder. Local occupancy values $\mathrm{occ}_l$ are then queried at predefined gripper control points (illustrated schematically in Figure \ref{fig:pipeline} (b)), which are divided into (i) outer gripper points ($\mathcal{O}^\text{outer}$) used for collision evaluation and (ii) inner gripper points ($\mathcal{O}^\text{inner}$) corresponding to the intended contact region between the fingers. See Appendix Figure \ref{fig:refine_points} for the full set of outer and inner gripper points. The occupancy predictions at those locations ($\hat{\mathcal{O}}^\text{outer}$, $\hat{\mathcal{O}}^\text{inner}$) are then used to determine the cost of the sampled perturbation using the objective: 
\vspace{-2mm}

\definecolor{collcolor}{RGB}{255, 151, 65}
\definecolor{contactcolor}{RGB}{255, 104, 149}
\definecolor{regcolor}{RGB}{137, 104, 246}

\begin{equation}
    \mathcal{J}=
    w_{\text{free}}
    \underbrace{
    \frac{1}{N_o}\sum_{i=1}^{N_o}\hat{o}_i^{\text{outer}}
    }_{\textcolor{collcolor}{\text{collision penalty}}}
    +
    w_{\text{contact}}
    \underbrace{
    \left(
    1-\min\!\big(
    \mathrm{TopK}(\hat{\mathcal{O}}^{\text{inner}},K)
    \big)
    \right)
    }_{\textcolor{contactcolor}{\text{contact encouragement}}}
    +
    w_{\text{reg}}\!\!\!\!\!\!
    \underbrace{
        \|\xi\|_2^2
    }_{\textcolor{regcolor}{\text{regularization}}}
\end{equation}

where $\hat{o}_i^{\text{outer}} \in \hat{\mathcal{O}}^{\text{outer}}$, $N_o = |\mathcal{O}^{\text{outer}}|$, and $K < |\mathcal{O}^{\text{inner}}|$. 
The collision term \textcolor{collcolor}{penalizes intersections} between the gripper and object, while the contact term 
\textcolor{contactcolor}{encourages stable contact} with the contact surface enclosed by the gripper and the regularization term maintains \textcolor{regcolor}{local updates}. Together, $\mathcal{J}$ characterizes locally valid grasps based on the surrounding local geometry.
We use an MPPI \cite{mppi} style weight update to iteratively minimize this objective, correcting local geometric errors while preserving the overall dual-arm grasp configuration by implicitly reasoning about the missing local geometry.

\begin{figure}
    \centering
    \includegraphics[width=0.9\linewidth]{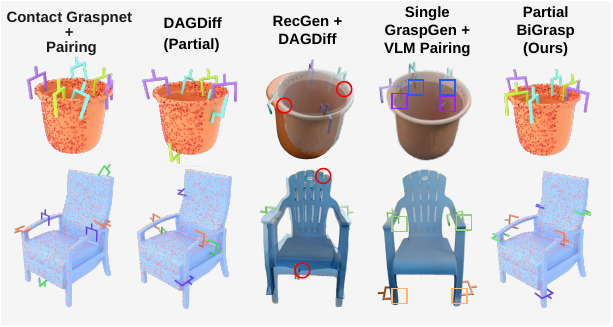}
    \vspace{-1mm}
    \caption{\small Qualitative comparison. ContactGraspNet + Pairing exhibits poorly aligned grasps and collisions, visible around the bucket rim and chair armrests, due to limited geometric context and unreliable grasp pairing. DAGDiff (Partial) mistakes artificial boundaries caused by occlusion for valid grasp surfaces, resulting in collisions on the bucket rim and chair leg. RecGen + DAGDiff (Full) suffers from reconstruction and registration errors, leading to colliding grasps on the bucket rim and chair backrest, as highlighted by the red circles. Single GraspGen + VLM Pairing predicts grasp regions inside the bucket, where valid external contacts cannot be established, and produces unstable grasp regions around the chair legs, as indicated by the coloured bounding boxes. In contrast, PartialBiGrasp (Ours) reasons jointly about global object structure and local grasp geometry to generate well-aligned, collision-free grasp pairs.}
    \label{fig:Qualitative}
\end{figure}

\section{Experimental Results}
\label{sec:experiments}

\subsection{Setup}

\paragraph{Training and Evaluation.} All models are trained on the DG16M dataset \cite{dg16m}. Training is performed in three stages. First, the Global Geometry Encoder together with the single-grasp synthesis heads are jointly trained using occupancy supervision, grasp pose supervision, and graspability region supervision derived from valid grasp contacts in DG16M. Next, the Dual-Grasp Critic is trained to classify whether candidate grasp pairs satisfy force-closure constraints. Finally, the Local Geometry Encoder is trained using local occupancy supervision to enable occupancy-guided grasp refinement under partial observations. 
The framework is evaluated in simulation on an unseen split of DG16M and a \textit{real-world dataset} comprising 11 objects, each captured from three viewpoints using a RealSense D455, along with corresponding \textit{digital-twin} meshes. Real RGB-D observations are provided as input to the methods, while the digital-twin meshes are used only to compute evaluation metrics in simulation.
To assess sim-to-real transfer, we additionally evaluate grasp execution using our model on six objects with a heterogeneous dual-arm setup consisting of an xArm7 and an xArm6 Lite, with motion planning performed using cuRobo~\cite{curobov2}. All training and inference runs are performed on a single NVIDIA RTX 3090 GPU.

\vspace{-2mm}
\paragraph{Baselines.} We compare our framework against representative approaches for grasp generation under partial observations. First, we evaluate ContactGraspNet~\cite{contact_graspnet} for single-arm grasp generation combined with two grasp pairing strategies: an attention-based pairing module and a classifier-based pairing network. Second, we train DAGDiff~\cite{dagdiff} directly on partial point cloud observations to evaluate diffusion-based dual-arm grasp generation under incomplete geometry, denoted by DAGDiff(Partial). Third, we evaluate a reconstruction-based pipeline using RecGen~\cite{recgen} followed by DAGDiff trained on completed object geometry, denoted by RecGen + DAGDiff(Full). Finally, we compare against a VLM-based grasp pairing baseline~\cite{taskawarebimanualaffordanceprediction} that predicts grasp-region compatibility from RGB semantic cues, given predicted single grasps from our model, denoted by Single GraspGen + VLM Pairing. The RecGen and the VLM-based baseline are evaluated only on the \textit{real-world dataset}, as they require RGB observations containing object details and photorealistic textures unavailable in rendered DG16M ShapeNet~\cite{shapenet} CAD models. All baselines are configured to generate 100 high-scoring grasp pairs per object for fair comparison and are evaluated by the following metrics.

\vspace{-2mm}
\paragraph{Metrics.}  We evaluate all models using Force Closure (FC\%), Grasp Success Rate (GS\%), Pair Collision Rate (PC\%), and Coverage. FC\% measures whether predicted grasp pairs satisfy analytical force-closure constraints described in \cite{dg16m}. GS\% evaluates the physical success of grasps in Isaac Gym~\cite{isaac_gym} by testing whether objects can be stably lifted and held under gravity using floating grippers. PC\% measures the percentage of predicted grasp pairs that intersect the object geometry, reflecting collision avoidance and geometric validity. Coverage measures how well the set of force-closure-valid grasps spans the graspable regions of the object surface.

\subsection{Results}

\begin{table*}
\centering
\caption{\small Quantitative comparison of grasp generation methods on DG16M and the real-world dataset.
}
\label{tab:main_results}

\vspace{-1mm}

\renewcommand{\arraystretch}{1}
\setlength{\tabcolsep}{5pt}

\resizebox{\textwidth}{!}{
\begin{tabular}{lccccccc}
\toprule

\multirow{2}{*}{\textbf{Method}}
& \multicolumn{4}{c}{\textbf{DG16M}}
& \multicolumn{3}{c}{\textbf{Real-World Dataset}} \\

\cmidrule(lr){2-5}
\cmidrule(lr){6-8}

& \textbf{FC\%$\uparrow$}
& \textbf{GS\%$\uparrow$}
& \textbf{PC\%$\downarrow$}
& \textbf{Cov.$\uparrow$}

& \textbf{FC\%$\uparrow$}
& \textbf{GS\%$\uparrow$}
& \textbf{PC\%$\downarrow$} \\

\midrule

ContactGraspNet \cite{contact_graspnet} + Attn Pairing
& 21.10
& 52.14
& 47.41
& 10.11
& 22.56
& 65.04
& 57.15 \\

\rowcolor{gray!10}
ContactGraspNet \cite{contact_graspnet} + Classifier Pairing
& 15.03
& 50.05
& 63.44
& 9.42
& 22.22
& 62.60
& 63.44 \\

DAGDiff \cite{dagdiff} (Partial)
& 22.35
& 46.63
& 49.19
& 14.48
& 17.47
& 62.47
& 59.27 \\

\rowcolor{gray!10}
RecGen \cite{recgen} + DAGDiff \cite{dagdiff} (Full)
& --
& --
& --
& --
& 20.51
& 64.33
& 53.2 \\

Single GraspGen + VLM Pairing~\cite{taskawarebimanualaffordanceprediction}
& --
& --
& --
& --
& 22.96
& 66.93
& 50.16 \\

\midrule

\rowcolor{gray!15}
\textbf{Ours}
& \textbf{55.16}
& \textbf{67.87}
& \textbf{17.48}
& \textbf{16.96}
& \textbf{51.06}
& \textbf{81.54}
& \textbf{38.80} \\

\bottomrule
\end{tabular}
}

\end{table*}
\begin{table}
\centering
\caption{\small Ablation study of different components of our method on DG16M and the real-world dataset.
}
\label{tab:ablation}

\vspace{1mm}

\renewcommand{\arraystretch}{1}
\setlength{\tabcolsep}{7.5pt}

\begin{tabular}{lcccc ccc}
\toprule

\multirow{2}{*}{\textbf{Method}}
& \multicolumn{4}{c}{\textbf{DG16M}}
& \multicolumn{3}{c}{\textbf{Real-World Dataset}} \\

\cmidrule(lr){2-5}
\cmidrule(lr){6-8}

& \textbf{FC\%$\uparrow$}
& \textbf{GS\%$\uparrow$}
& \textbf{PC\%$\downarrow$}
& \textbf{Cov.$\uparrow$}

& \textbf{FC\%$\uparrow$}
& \textbf{GS\%$\uparrow$}
& \textbf{PC\%$\downarrow$} \\

\midrule

No Refinement
& 45.35
& 61.51
& 32.61
& 15.09
& 40.74
& 76.47
& 44.68 \\

\rowcolor{gray!10}
Global Refinement
& 39.57
& 63.67
& 35.30
& 14.42
& 40.85
& 74.40
& 42.74 \\

No Residual
& 19.85
& 41.70
& 67.63
& 7.15
& 27.02
& 65.31
& 48.50 \\

\midrule

\rowcolor{gray!15}
\textbf{Ours}
& \textbf{55.16}
& \textbf{67.87}
& \textbf{17.48}
& \textbf{16.96}
& \textbf{51.06}
& \textbf{81.54}
& \textbf{38.80} \\

\bottomrule
\end{tabular}

\end{table}

\vspace{-2mm}
\textbf{Baseline Comparison.} 
Table~\ref{tab:main_results} reports quantitative results, and Figure~\ref{fig:Qualitative} shows qualitative comparisons with each baseline. ContactGraspNet achieves relatively low FC\% and GS\% due to its inability to represent a region on the object with a context rich feature; it simply inherits the feature of the nearest point. Its reliance on dense point-wise grasp supervision further limits performance for large objects with sparse graspable regions, where grasp quality degrades beyond the few high-confidence grasps, resulting in high PC\% and lower coverage. The attention-based pairing method, which learns grasp pairing directly from point features of single grasps through online supervision along the lines of~\cite{bigraspformer}, performs marginally better than the classifier-based pairing network trained on DG16M grasp annotations, as the latter relies on nearest-point descriptors for grasp pairing, which may not adequately capture the geometric context of the grasp candidates. 

DAGDiff is designed for complete object geometry and therefore exhibits a substantial performance drop when trained directly on partial point clouds. In particular, it frequently assigns grasps to artificial boundaries introduced by occlusions and incomplete observations, leading to a high collision rate, as illustrated in Figure \ref{fig:Qualitative}. Since collision refinement is performed on the observed geometry, missing contact surfaces further limit its ability to reject invalid grasp configurations.

The RecGen + DAGDiff pipeline inherits errors from both reconstruction and registration stages. Reconstruction artifacts and pose inaccuracies can introduce grasp offsets that produce unstable or colliding grasp pairs, reducing FC\% and GS\% while increasing PC\%. Reconstructing dense geometry also introduces significant computational overhead. For generating 100 grasp pairs, our method requires \(3.32 \pm 0.87\) s, compared to \(20.81 \pm 0.27\) s for DAGDiff and \(36.39 \pm 1.54\) s for RecGen + DAGDiff, with corresponding mean peak allocated GPU memory of \(6.68\), \(1.52\), and \(10.15\) GB, respectively. Reducing the number of generated pairs to 50 decreases our inference time by approximately \(20\%\), while the baseline methods' runtimes remain nearly unchanged.

The VLM-based method reasons over semantic object regions rather than explicit geometric constraints. While this often identifies intuitively opposing grasp locations, it does not directly model force-closure stability, contact geometry, or collision constraints, resulting in lower FC\% and reduced overall grasp quality.

In contrast, our method directly learns grasp generation, force-closure reasoning, and local geometric understanding within a unified framework. The continuous convolutional representation provides rich geometric features for both single-grasp prediction and grasp-pair evaluation, and query-based supervision naturally handles sparse graspable regions without requiring dense per-point labels. The Force Closure Critic learns physically grounded grasp-pair compatibility from DG16M, enabling reliable discrimination between stable and unstable grasp configurations. To address ambiguities introduced by partial observations, graspability prediction identifies promising contact regions, while the Local Encoder predicts local occupancy around each grasp candidate. By operating directly on occupancy queries rather than reconstructing full surfaces, our approach remains computationally efficient while achieving substantially higher FC\%, GS\%, and coverage, together with significantly lower collision rates, on both DG16M and noisy observations of novel objects from the real-world dataset.

\textbf{Ablation Studies.}
Table \ref{tab:ablation} evaluates the contribution of the key components in our framework. 
In \textit{Global Refinement}, the refinement objective is evaluated using occupancy predicted by the Global Encoder instead of the Local Encoder. This variant leads to lower FC\% and higher collision rates than our method, indicating that coarse object-level geometry alone is insufficient for accurately reasoning about grasp-contact regions under partial observations. Introducing local occupancy-guided refinement substantially improves grasp stability and reduces collisions over \textit{No Refinement} by incorporating fine-grained geometric information around candidate contact regions. Although force closure is not explicitly optimized during refinement, FC\% still increases, as correcting contact placement and removing gripper-object collisions allows grasp pairs that previously failed the force-closure criterion to satisfy it.

Finally, removing the residual connection (\textit{No Residual}) between global and local features causes a significant performance degradation across all metrics. Under partial visibility, a local crop is often ambiguous, as the same visible patch may correspond to
different objects and different hidden geometry. The residual connection provides an object-level context to resolve these ambiguities; without it, less reliable occupancy predictions lead to poorer refinement and higher collision rates.
Together, these results demonstrate that both local refinement and the fusion of global and local representations are critical to the performance of our method.

\begin{wrapfigure}{r}{0.45\columnwidth}
    \centering
    \vspace{-10pt}
    \includegraphics[width=0.45\columnwidth]{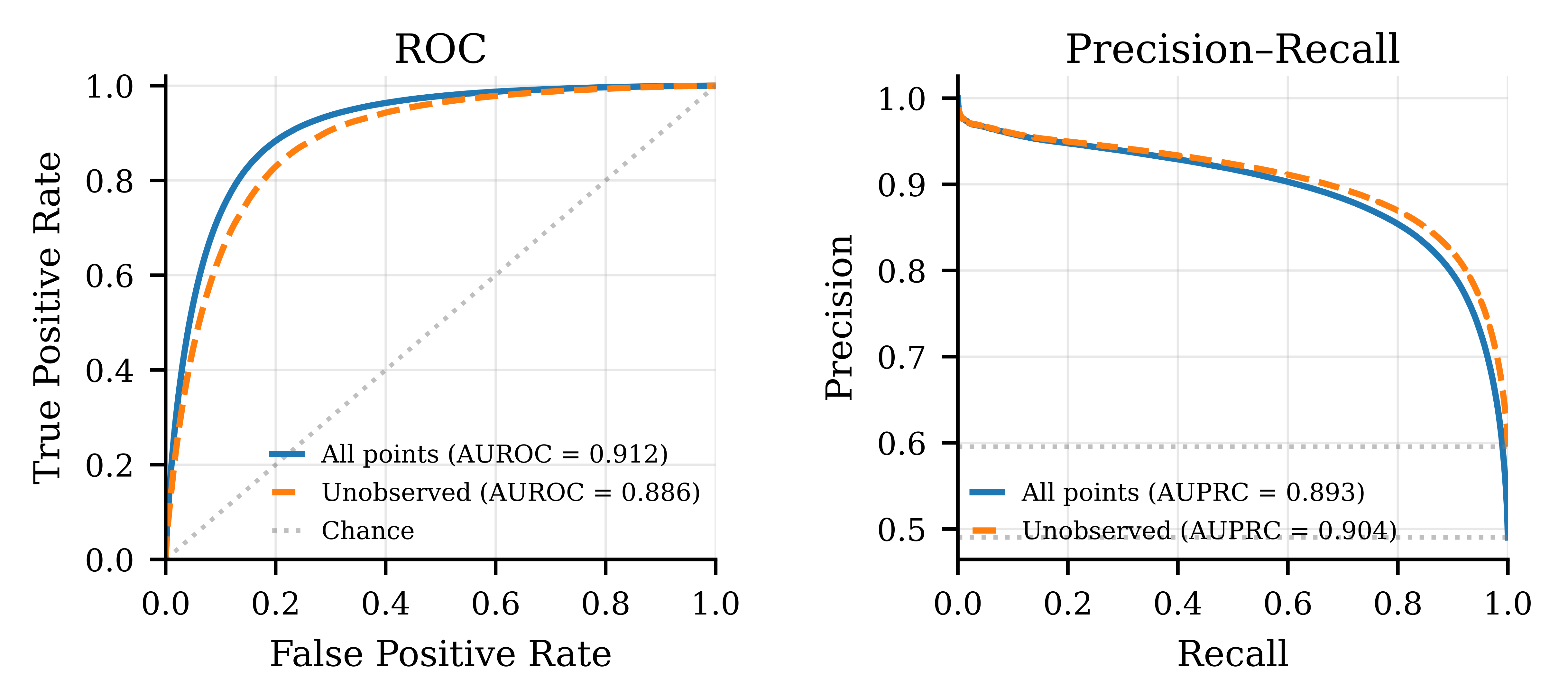}
    \caption{\small Occupancy prediction performance on the DG16M test set. Left: ROC curves. Right: precision recall curves. Solid blue curves denote predictions across all query points, while dashed orange curves denote predictions restricted to unobserved regions.}
    \vspace{-14pt}
    \label{fig:roc_pr}
\end{wrapfigure}

\textbf{Geometry Representation Evaluation. }
To quantitatively evaluate the geometric information captured by our learned representation, we assess the Local Encoder's occupancy predictions against ground-truth occupancy independently of grasp success. We sample 500 query points around each of 64 grasps per object from the DG16M test set and evaluate the predicted occupancy probabilities using AUROC and AUPRC. As shown in Figure \ref{fig:roc_pr}, our model achieves an AUROC of 0.912 and an AUPRC of 0.893 across all query points (\textcolor[HTML]{1F77B4}{\rule[2pt]{12pt}{1pt}} curve), indicating strong agreement with ground-truth occupancy. We further evaluate predictions restricted to unobserved regions (\textcolor{orange}{\rule[2pt]{3pt}{1pt}\rule[2pt]{1.5pt}{0pt}\rule[2pt]{3pt}{1pt}\rule[2pt]{1.5pt}{0pt}\rule[2pt]{3pt}{1pt}} curve), where distinguishing hidden occupied space from free space is particularly challenging under partial observations. Despite the absence of direct geometric observations in these regions, the model achieves an AUROC of 0.886 and an AUPRC of 0.904. These results indicate that the Local Encoder learns a meaningful geometric representation rather than merely fitting the observed point cloud. Together with the ablation results, which show that local occupancy-guided refinement improves grasp quality and reduces collisions, they support the role of accurate local geometry prediction in enabling more reliable grasp refinement under partial observations.

\begin{figure}
    \centering
    \includegraphics[width=1\linewidth]{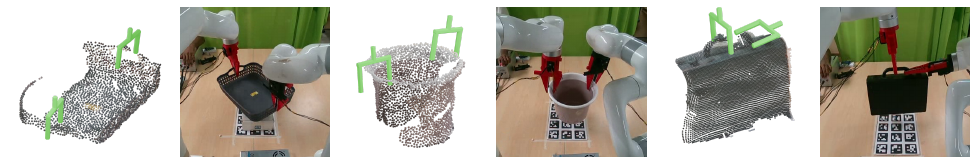}
    \caption{\small Real-World Execution. For six objects from the \textit{Real-World Dataset}, we also execute grasps on a heterogeneous dual-arm setup consisting of an xArm7 and xArm6 Lite. Note: The briefcase has no grasps on the sides because it is thicker than the maximum gripper opening.}
    \label{fig:realdemo}
    \vspace{-14pt}
\end{figure}

\textbf{Real-World Execution.} On real-world robot arms and noisy point clouds (Figure \ref{fig:realdemo}), we demonstrate successful transfer of our method with an average success rate of $\sim$72\% across six objects as seen in Table \ref{tab:real_robot_success}. For each object, we execute the top three reachable pairs from each of three views, resulting in nine trials per object. 

\begin{table}[t]
    \centering
    \caption{\small Real-robot grasp success rates across six objects.}
    \label{tab:real_robot_success}
    \vspace{1mm}
    \renewcommand{\arraystretch}{1}
    \setlength{\tabcolsep}{13pt}
    \begin{tabular}{lcccccc}
        \toprule
        \textbf{Object} & Chair & Stool & Bucket & Briefcase & Saucepan & Tray \\
        \midrule
        \textbf{Success} & 7/9 & 7/9 & 8/9 & 6/9 & 5/9 & 6/9 \\
        \bottomrule
    \end{tabular}
    \vspace{-10pt}
\end{table}

\section{Conclusion}
\label{sec:conclusion}
In this paper, we present PartialBiGrasp, an occupancy-conditioned framework for generating stable dual-arm grasp pairs from partial point clouds. Our framework combines a global occupancy encoder that reasons about object geometry, graspability, and dual-arm grasp pairing with a grasp-aligned local occupancy encoder that refines individual grasp candidates using local surface geometry. The global encoder learns a continuous object-level representation that captures object structure from sparse observations and enables robust reasoning over the limited graspable regions where valid grasps exist. The local encoder, conditioned on the global features, focuses on the geometry around each grasp candidate and refines grasp poses through a lightweight sampling-based optimization procedure. This hierarchical design enables efficient reasoning at both object and contact scales without requiring explicit reconstruction at inference time. Experimental results on both synthetic and noisy real-world partial point clouds demonstrate improved force-closure rates, higher grasp simulation success rates during object lifting, and fewer collisions compared to existing baselines.

\section{Limitations}
\label{sec:limitations}
Our framework does not explicitly model grasp reachability or motion feasibility for a given embodiment during grasp generation. As a result, some predicted grasp pairs may be infeasible on dual-arm systems due to kinematic constraints or arm collisions, requiring a motion planner to validate and execute grasps for deployment. Additionally, occupancy predictions rely on learned geometric priors and may benefit from explicit uncertainty estimation to better handle ambiguous out-of-distribution geometry.

\acknowledgments{This work was funded in part by grants made available via DIA-CoE, IIT Delhi. The authors thank the anonymous reviewers and the Area Chair for their constructive feedback.}

\clearpage

\bibliography{cite}  %

\clearpage

\appendix
\renewcommand{\thesubsection}{\Alph{subsection}}

\section*{Appendix}

\subsection{Dataset}
\label{sec:Dataset}
\subsubsection{Dual Arm Grasping Dataset}
\label{subsec:Dataset}
We train and evaluate our method using the DG16M dataset, which contains 4,142 large-object meshes from ShapeNet annotated with approximately 2,000 positive and negative dual-arm grasp pairs per object generated using analytical force-closure evaluation. We reserve 300 objects for testing and use the remaining objects for training. For each object, single-view partial observations are generated by rendering RGB-D images from randomly sampled viewpoints and converting the visible depth measurements masked by the object into partial point clouds. All quantitative results are reported on this unseen test split.

\begin{wrapfigure}{r}{0.35\textwidth}
    \centering
    \vspace{-30pt}
    \includegraphics[width=0.32\textwidth]{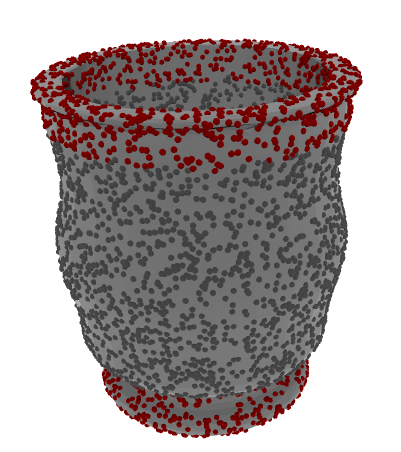}
    \caption{Graspability map for a vase. Red points indicate high-graspability regions, while gray points indicate lower-graspability regions.}
    \label{fig:graspness}
    \vspace{-10pt}
\end{wrapfigure}

\paragraph{Single-Grasp and Graspability Label Generation.}
To train the single-grasp generation module, we extract unique valid single-arm grasps from the grasp pairs provided in DG16M. The resulting grasp poses and contact locations are used as supervision for grasp pose regression. Graspability labels are derived from these contact locations, where positive samples correspond to valid single-grasp contacts, and negative samples are drawn from regions assigned zero graspability score by an annotated graspability map, as shown in Figure \ref{fig:graspness}. It is constructed by propagating graspability scores from valid single-grasp contact locations from DG16M data using a distance-based weighting function over the object surface.

\paragraph{Visibility Filtering.}

Since training is performed on single-view partial observations, only grasp annotations whose contact locations are visible in the current observation are used for supervision. For dual-arm grasp pairs, both contact locations must be visible for the pair to be retained. This visibility filtering is applied consistently across all methods trained on partial observations of DG16M objects.

\paragraph{Force-Closure Grasp Pair Labels.}
The Dual-Grasp Critic is trained using force-closure labels obtained from DG16M. Positive grasp pairs correspond to analytically verified force-closure grasps provided by the dataset, while negative pairs are sampled from non-force-closure grasp combinations. These labels are used to train the critic as a binary classifier that predicts whether a candidate grasp pair satisfies force-closure constraints.

\subsubsection{Occupancy Dataset}
To train the global and local convolutional occupancy encoders, we generate occupancy supervision from the watertight object meshes. Query points are sampled both near the object surface and throughout the object bounding volume, and assigned binary occupancy labels. Global occupancy labels supervise object-scale geometric reasoning, while local occupancy labels are generated from grasp-centered local regions to train the Local Geometry Encoder.

\subsubsection{Real-World Dataset}
\begin{figure}[t]
    \centering

    \begin{minipage}{0.55\linewidth}
        \centering
        \includegraphics[width=\linewidth]{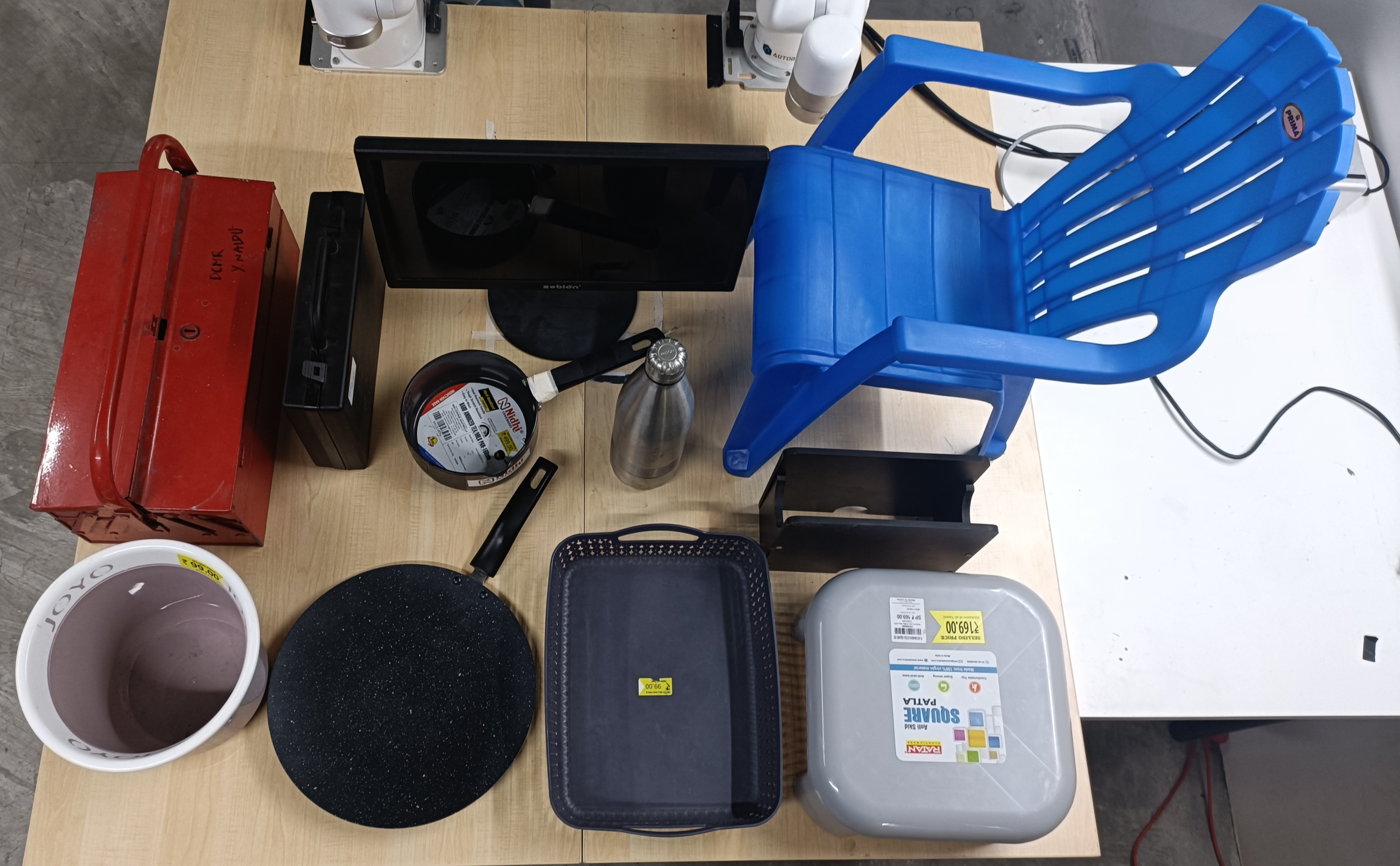}
        \captionof{figure}{Real-world object dataset.}
        \label{fig:real_world}
    \end{minipage}
    \hfill
    \begin{minipage}{0.38\linewidth}
        \centering
        \includegraphics[width=\linewidth]{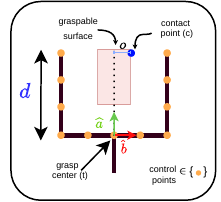}
        \captionof{figure}{Y-shaped gripper used in our experiments.}
        \label{fig:gripper}
    \end{minipage}

\end{figure}

We collect a dataset of 11 large everyday objects captured using an Intel RealSense D455 RGB-D camera, as shown in Figure \ref{fig:real_world}. Each object is observed from three viewpoints and represented as a partial point cloud obtained from the depth image. The dataset includes chairs, stools, buckets, monitors, containers, toolboxes, and other objects with diverse geometry and graspable structures. Digital twin meshes are created for evaluation in simulation. %

\subsection{Architecture and Training Details}
\subsubsection{Global Encoder}
Given a partial point cloud observation \(\mathcal{P}\in \mathbb{R}^{2048\times3}\), we employ the Global Geometry Encoder to extract an occupancy-conditioned representation of the object. The encoder uses a VN-PointNet encoder to compute rotation-equivariant point features, which are projected onto three orthogonal feature planes and processed using a shared 2D U-Net, as shown in Figure \ref{fig:triplane_comparison}. Features at arbitrary 3D locations are obtained through bilinear interpolation, yielding a continuous occupancy-conditioned feature field. 

\paragraph{Grasp Generation Module.}
For each candidate contact location, we define a fixed set of \(N_{query}=30\) query points arranged within a spherical neighborhood centered at the contact point. This canonical query set provides a consistent local geometric representation that can be evaluated at arbitrary contact locations through the continuous occupancy field. Occupancy-conditioned features extracted at the query locations are projected by the Grasp Generation Module \(\mathcal{G}\) to a lower-dimensional latent representation. Occupancy predictions are produced independently at each query location, while the latent features from all query points are concatenated to form a grasp descriptor used by the graspability and grasp regression heads.

The grasp descriptor is processed by multiple prediction heads to predict a graspability score \({p}_G\), grasp approach direction \(\hat{\mathbf a}\), baseline direction \(\hat{\mathbf b}\), and gripper offset \({o}\). The predicted approach and baseline vectors are first orthonormalized using the Gram-Schmidt method, after which the grasp rotation and translation are reconstructed as

\begin{equation}
R =
\begin{bmatrix}
\hat{\mathbf b} &
\hat{\mathbf a} &
\hat{\mathbf b}\times\hat{\mathbf a}
\end{bmatrix},
\qquad
\mathbf t =
\mathbf c - o\,\hat{\mathbf b} - d\,\hat{\mathbf a},
\end{equation}

where \(\mathbf c\) denotes the contact point and \(d\) is the gripper depth as shown in Figure \ref{fig:gripper}. The resulting grasp pose is represented as a rigid transformation \(G=(R,\mathbf t)\in SE(3)\). Throughout, $(\cdot)^{*}$ denotes ground truth; symbols without it denote network predictions.
The occupancy and grasp prediction heads are trained jointly using

\begin{equation}
    \mathcal{L}_{G}
    =
    \lambda_{occ}\,\mathcal{L}_{occ}
    +
    \lambda_{g}\,\mathcal{L}_{graspability}
    +
    \lambda_{cp}\,\mathcal{L}_{cp}
    +
    \lambda_{off}\,\mathcal{L}_{offset}
\end{equation}

where,

\begin{equation}
    \mathcal{L}_{occ}
    =
    \mathrm{BCE}\big(\mathrm{occ}_g,\, \mathrm{occ^{*}}_g\big),
    \qquad
    \mathcal{L}_{graspability}
    =
    \mathrm{BCE}\big(p_G,\, p_G^{*}\big),
\end{equation}

\begin{equation}
    \mathcal{L}_{cp}
    =
    \frac{1}{N_c}
    \sum_{i=1}^{N_c}
    \left\|
    G\mathbf{x}_i
    -
    G^{*}\mathbf{x}_i
    \right\|_1,
    \qquad
    \mathcal{L}_{offset}
    =
    \left|o - o^{*}\right|,
\end{equation}

where \(\mathbf{x}_i\) denotes the \(N_c\) predefined gripper control points shown in Figure~\ref{fig:gripper}, and $G\mathbf{x}_i = R\mathbf{x}_i + \mathbf{t}$ (likewise for $G^{*}$). The control-point and offset losses are computed only at contact locations associated with valid positive single grasp annotations. To account for the sparsity of positive grasp-contact annotations in the dual-arm datasets, graspability supervision is computed using a balanced set of positive and negative samples.

\paragraph{Grasp Pairing.}
The Dual-Grasp Critic is trained using the positive and negative force-closure grasp pairs provided in the DG16M dataset. For each grasp in a pair, Global Geometry Encoder features are projected to a latent representation using the force-closure module \(FC\). The resulting pair descriptors are concatenated and used to predict whether the grasp pair satisfies force-closure constraints. To encourage order invariance, the two grasp descriptors are randomly swapped during training with probability \(0.5\). The critic is trained as a binary classifier using

\begin{equation}
    \mathcal{L}_{FC}
    =
    \mathrm{BCE}(p_{fc}, p^{*}_{fc}),
\end{equation}

where \(p_{fc}\) and \(p^{*}_{fc}\) denote the predicted and ground-truth force-closure labels, respectively.

\subsubsection{Local Encoder}

\begin{figure}[t]
    \centering

    \begin{minipage}[c]{0.28\textwidth}
        \centering
        \includegraphics[width=\linewidth]{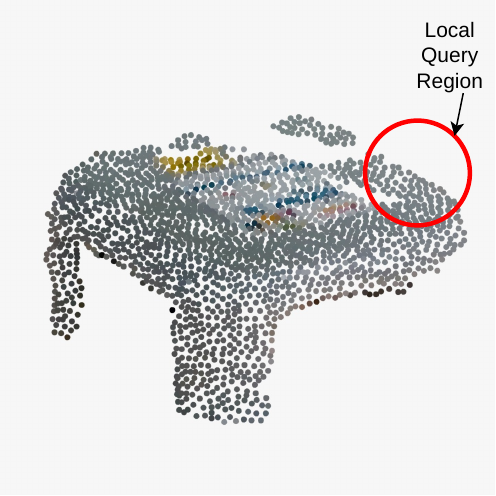}

        Input Point Cloud
    \end{minipage}
    \hfill
    \begin{minipage}[c]{0.68\textwidth}
        \centering

        \includegraphics[width=0.45\linewidth]{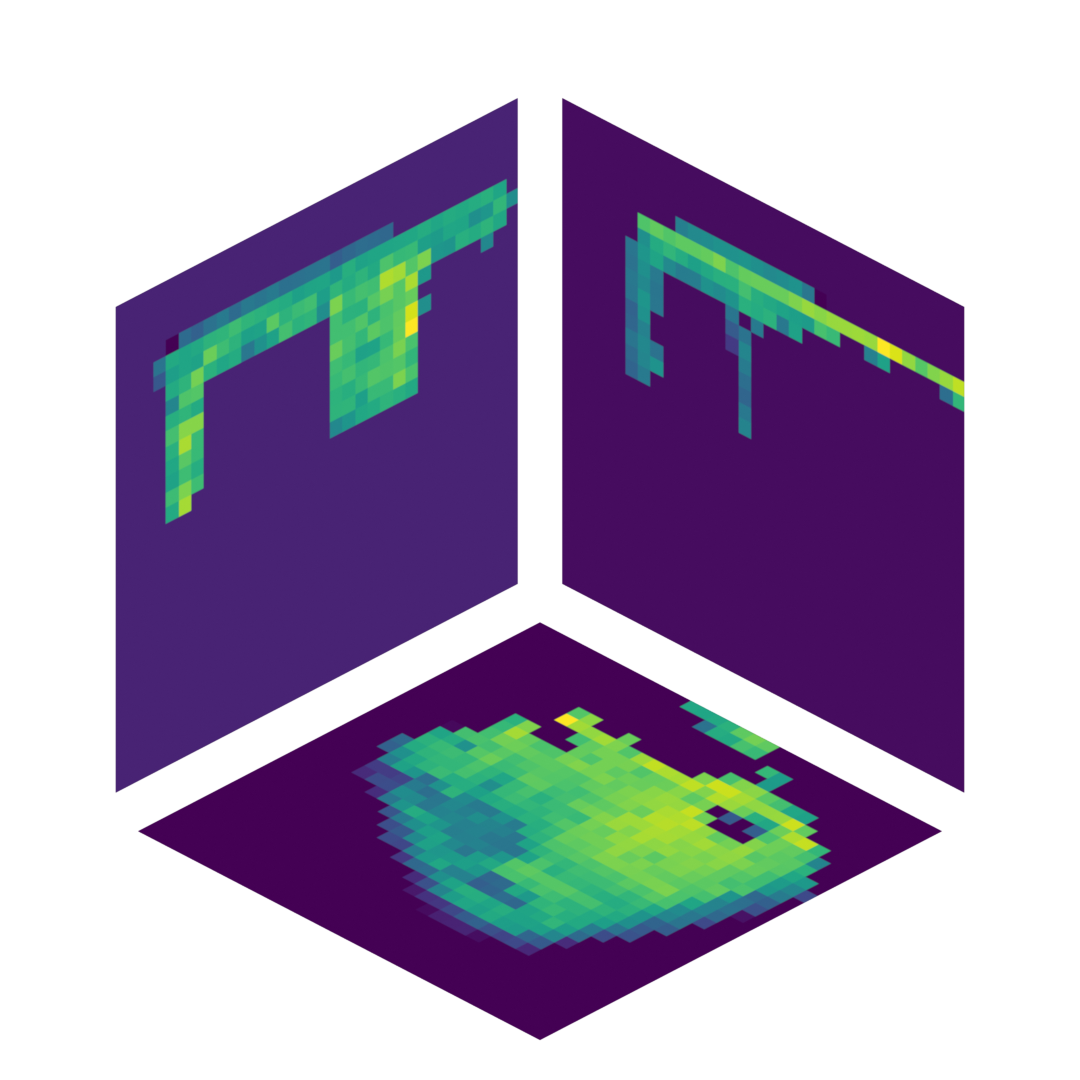}
        \includegraphics[width=0.45\linewidth]{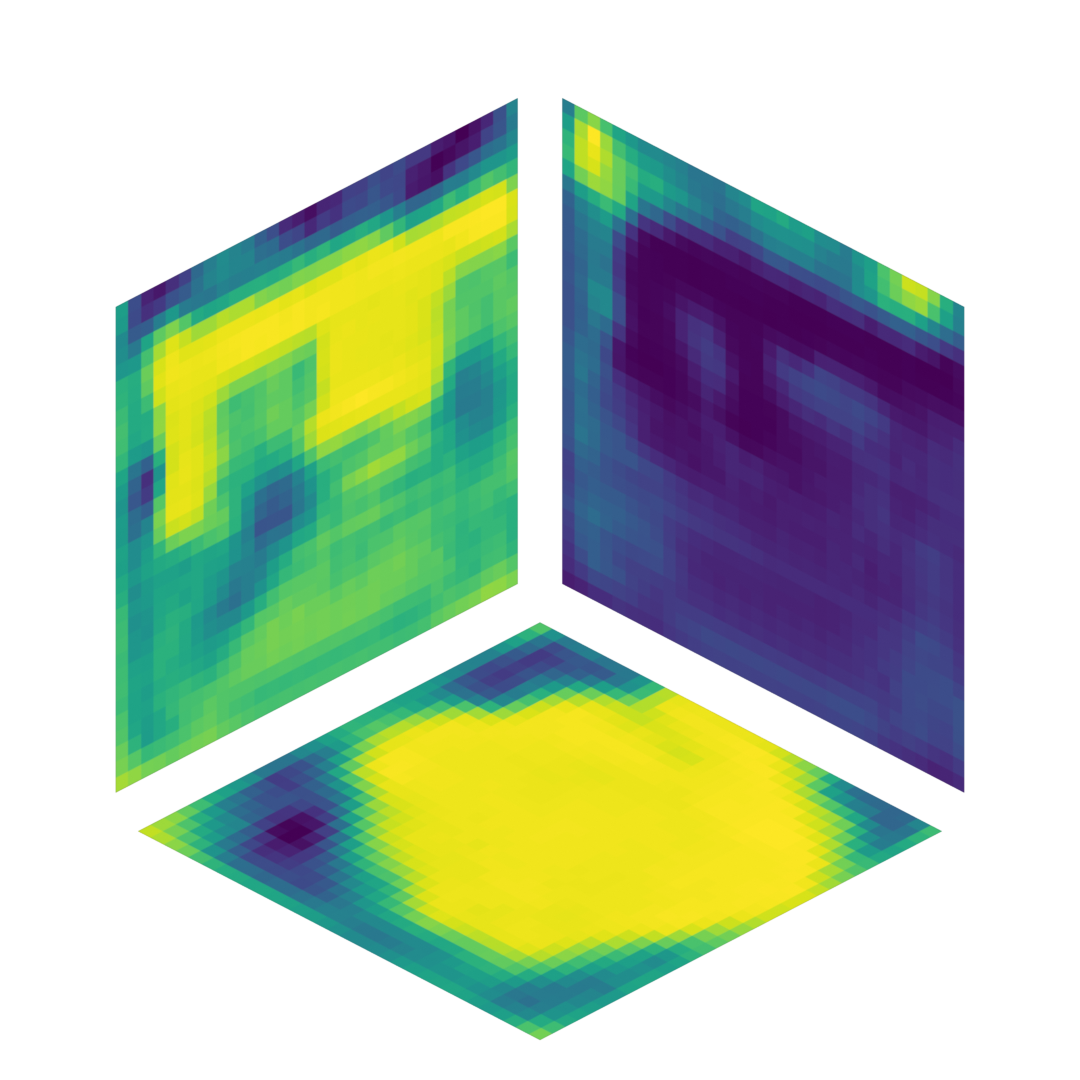}

        \small (A) Global Triplane

        \vspace{2mm}

        \includegraphics[width=0.45\linewidth]{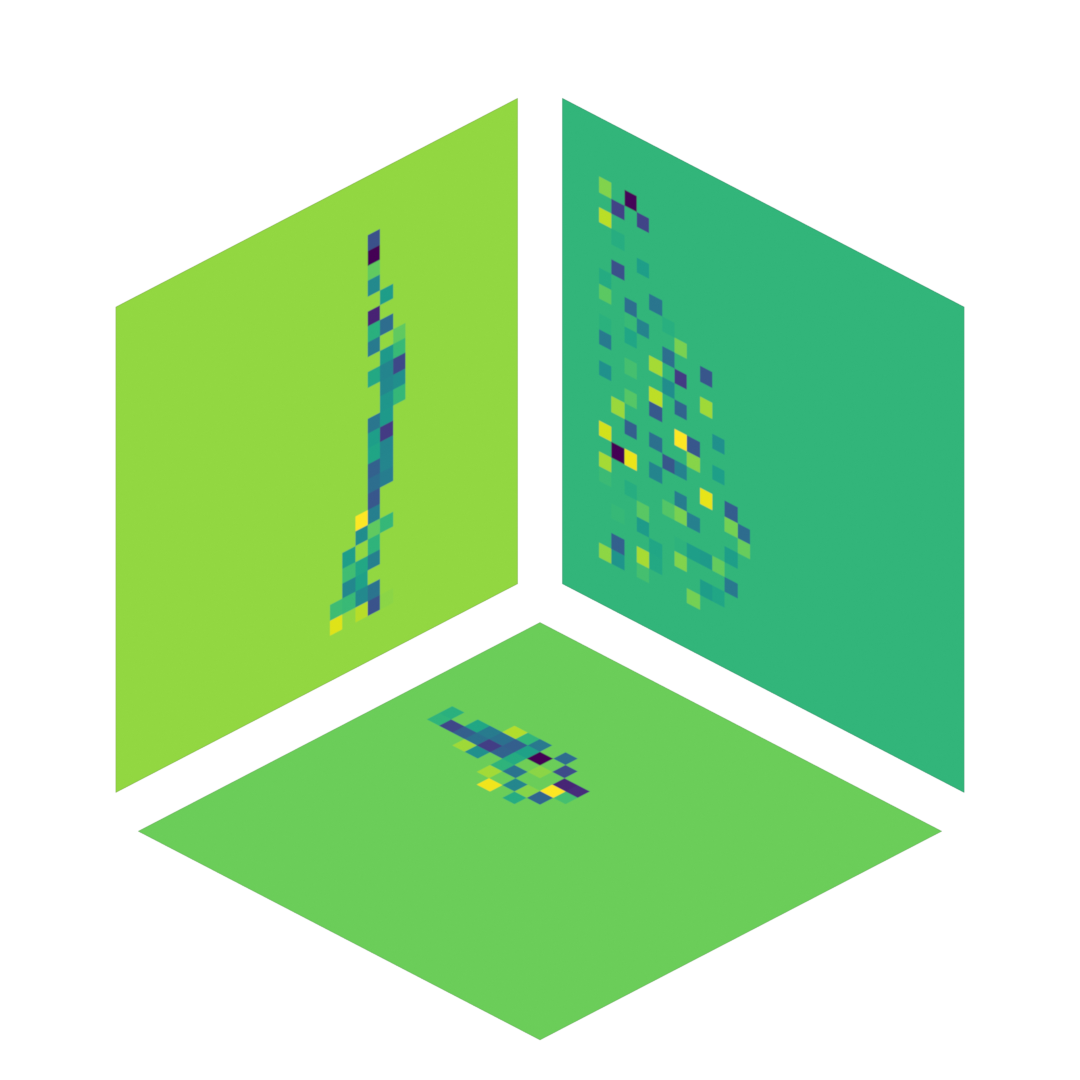}
        \includegraphics[width=0.45\linewidth]{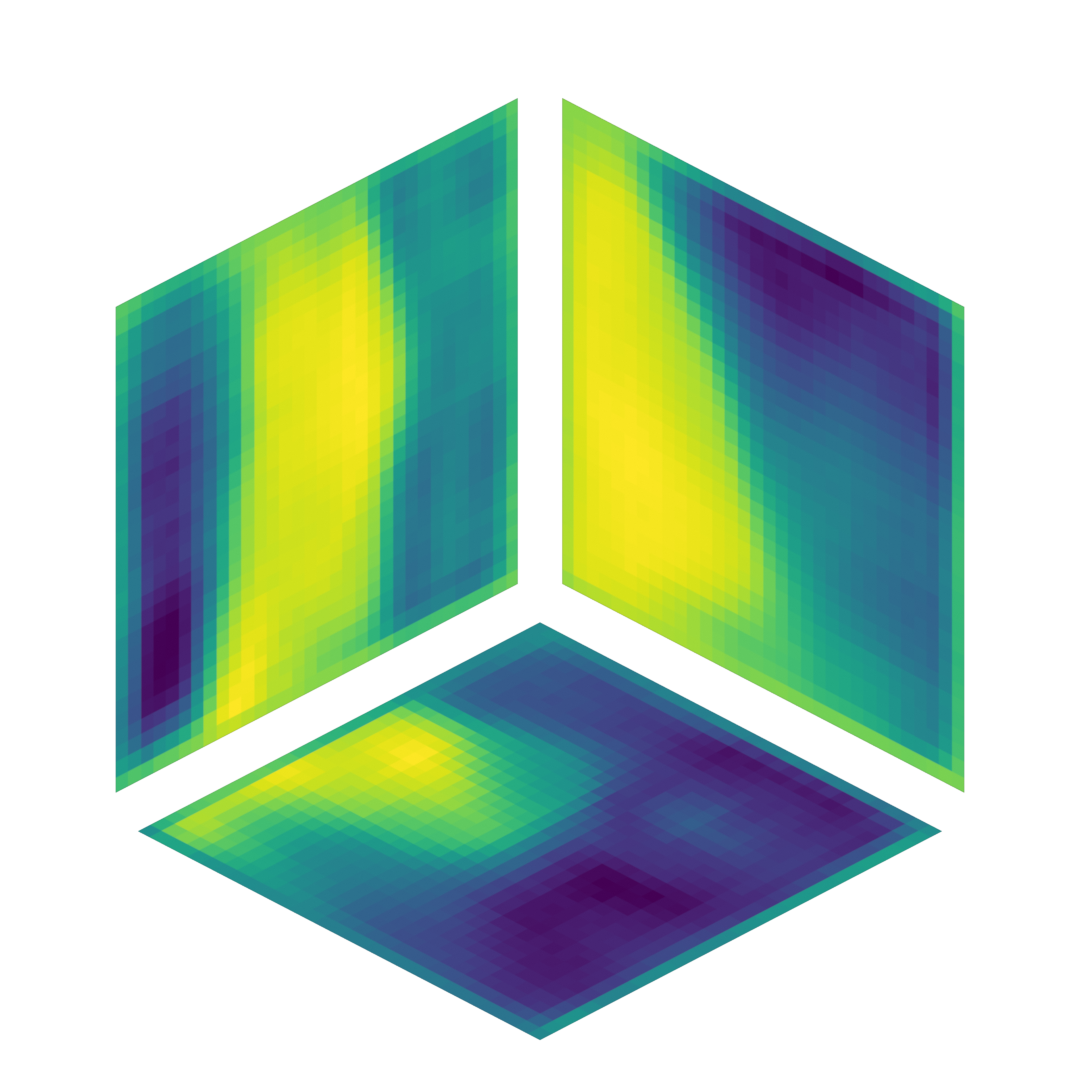}

        \small (B) Local Triplane
    \end{minipage}

    \caption{
    Triplane representations extracted from the input point cloud. For both (A) Global and (B) Local Triplanes, the left panel shows the initial triplane features and the right panel shows the features after U-Net processing.
    }
    \label{fig:triplane_comparison}
\end{figure}

\begin{wrapfigure}{r}{0.4\textwidth}
    \centering
    \includegraphics[width=0.36\textwidth]{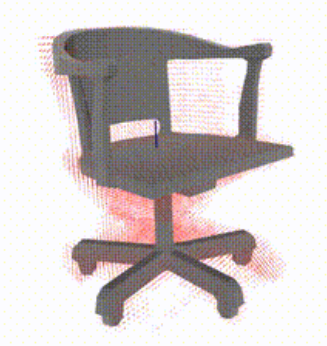}
    \caption{Coarse global occupancy output from the Global Encoder.}
    \label{fig:global_occ}
\end{wrapfigure}

The Local Encoder learns a geometry representation centered around a candidate grasp pose and is used both for local occupancy prediction and grasp refinement. Unlike the Global Encoder, which encodes the entire observed point cloud, the Local Encoder focuses exclusively on the neighborhood surrounding a grasp contact region. This helps us have a more fine grained representation of geometry that the Global Encoder does not provide (Figure \ref{fig:global_occ}).

Given a grasp pose $G \in SE(3)$ and a partial point cloud observation $\mathcal{P}=\{\mathbf{p}_i\}_{i=1}^{n}$, where $\mathbf{p}_i\in\mathbb{R}^3$, we first extract a local crop aligned with the grasp frame. Let $R_G \in SO(3)$ and $\mathbf{t}_G \in \mathbb{R}^3$ denote the rotational and translational components of the grasp pose, respectively. The crop center is defined as

\begin{equation}
\mathbf{m} = \mathbf{t}_G + \delta\,\mathbf{a}_G,
\end{equation}

where $\mathbf{a}_G$ denotes the grasp approach direction expressed in the world frame and $\delta$ is a fixed offset used to position the crop slightly ahead of the gripper along the approach axis.

Each point is transformed into the grasp coordinate frame centered at $\mathbf{m}$ according to

\begin{equation}
\tilde{\mathbf{p}}_i = R_G^\top (\mathbf{p}_i - \mathbf{m}).
\end{equation}

The local point cloud $\mathcal{P}_G$ is then defined as all points lying within an axis-aligned cube of side length $s$ and centered at $\mathbf{m}$,

\begin{equation}
\mathcal{P}_G
=
\left\{
\mathbf{p}_i \in \mathcal{P}
\;\middle|\;
\|\tilde{\mathbf{p}}_i\|_\infty \le \frac{s}{2}
\right\}.
\end{equation}

This local crop isolates the geometry most relevant to grasp quality while discarding distant object regions that are unlikely to influence contact formation.

Let $T_{\text{local}}^{\text{pre}}$ denote the local triplane features before local feature aggregation and let $T_{\text{global}}^{\text{post}}$ denote the globally aggregated triplane features obtained after the global U-Net. Global context is injected into the local representation through a residual connection,

\begin{equation}
T_{\text{local}}^{\text{res}}
=
T_{\text{local}}^{\text{pre}}
+
T_{\text{global}}^{\text{post}}.
\end{equation}

The resulting feature planes are then processed by the local U-Net as shown in Figure \ref{fig:triplane_comparison},

\begin{equation}
T_{\text{local}}^{\text{post}}
=
U_{\text{local}}
\!\left(
T_{\text{local}}^{\text{res}}
\right),
\end{equation}

where $U_{\text{local}}(\cdot)$ denotes the local feature aggregation network. This design enables the local representation to leverage object-level geometric context from the global branch while preserving fine-grained geometric information in the vicinity of the grasp.

Given a query point $q\in\mathbb{R}^3$, local features are interpolated from the triplane representation and passed through an occupancy decoder to predict the probability that the queried point lies inside the object surface.

The Local Encoder is supervised using the occupancy of local query points using

\begin{equation}
    \mathcal{L}_{occ_l} = \mathrm{BCE}\big(\mathrm{occ}_l,\, \mathrm{occ}^{*}_l\big)
\end{equation}

where $\mathrm{occ}_l$ and $\mathrm{occ}^{*}_l$ are the predicted and ground-truth occupancy labels respectively.

\begin{figure}[!t]
    \centering
    \includegraphics[width=1\linewidth]{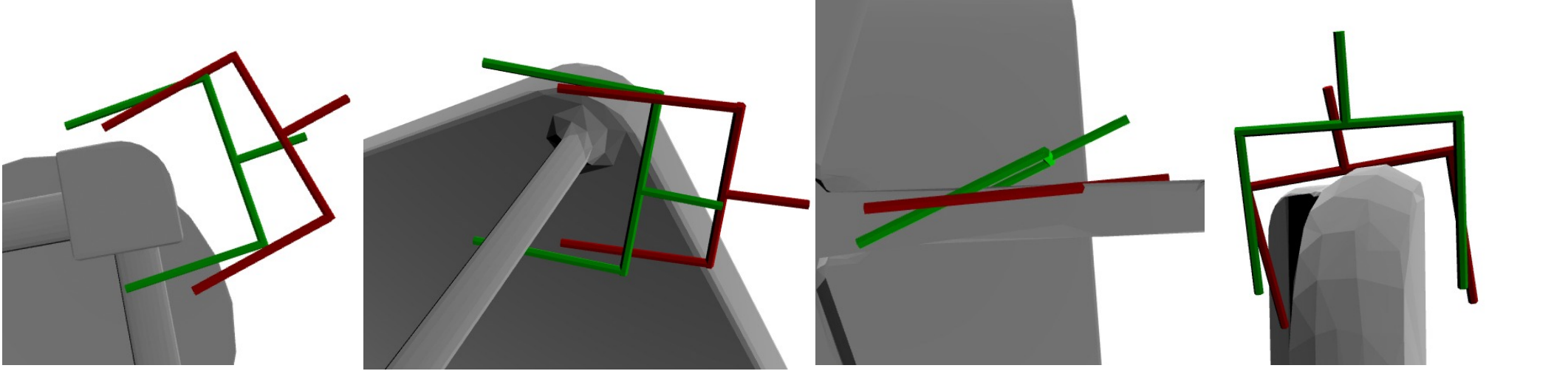}
    \caption{Qualitative examples of occupancy-guided grasp refinement. The original network prediction is shown in red, while the final refined grasp is shown in green. Refinement improves alignment with the object geometry   and yields more stable grasp configurations.}  
    \label{fig:refinement_exampples}
\end{figure}

\paragraph{Sampling-Based Occupancy-Guided Grasp Refinement.}

Although the grasp synthesis stage generates force-closure stable dual-arm grasp pairs, local geometric inconsistencies may still arise due to partial observations. %
To address these issues, we perform a lightweight occupancy-guided refinement procedure using the Local Encoder (Figure \ref{fig:refinement_exampples}).

Given the predicted dual-arm grasp pairs, we first extract the unique set of single-arm grasps and refine each independently. Refinement is formulated as an optimization problem in the Lie algebra $\mathfrak{se}(3)$, where candidate perturbations are applied to the current grasp pose and evaluated using local occupancy predictions.

Importantly, optimization is restricted to a small neighborhood around each predicted grasp. Consequently, refinement acts as a local geometric correction procedure that improves collision avoidance and contact quality while preserving the overall configuration predicted by the grasp pair synthesis stage.

Each perturbation is parameterized as

\begin{equation}
\xi
=
\bigg[ \Delta t_x,\Delta t_y,\Delta t_z,
\Delta r_x,\Delta r_y,\Delta r_z \bigg]
\in
\mathfrak{se}(3),
\end{equation}

and converted into a rigid-body transformation through the exponential map,

\begin{equation}
\label{eq:exp}
G' = G\,\exp(\xi).
\end{equation}

For each candidate grasp pose, a local crop is extracted and encoded using the Local Encoder. 

\begin{wrapfigure}{r}{0.45\textwidth}
    \centering
    \vspace{-30pt}
    \includegraphics[width=0.43\textwidth]{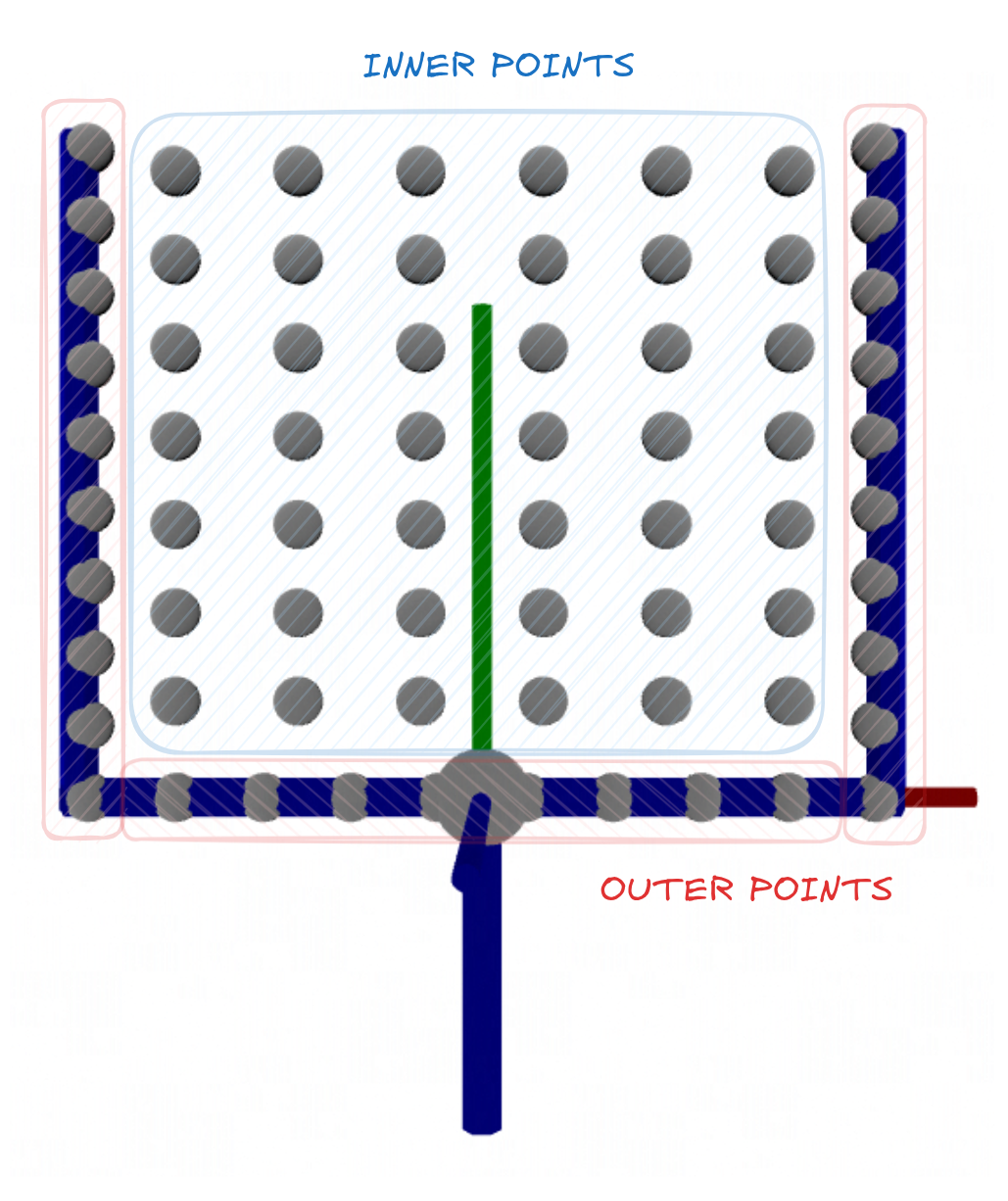}
    \caption{Control points used for occupancy-guided grasp refinement. Outer control points are used to evaluate collisions with the object, while inner control points are used to assess contact quality.}
    \label{fig:refine_points}
    \vspace{5pt}
\end{wrapfigure}

Occupancy values are queried at predefined gripper control points corresponding to the exterior gripper geometry and the finger closing region (Figure~\ref{fig:refine_points}).

Let

\begin{align}
\mathcal{O}^{\text{outer}} &= \left\{\mathbf{x}_i^{\text{outer}}\right\}_{i=1}^{N_o}, \\
\mathcal{O}^{\text{inner}} &= \left\{\mathbf{x}_i^{\text{inner}}\right\}_{i=1}^{N_{\text{in}}},
\end{align}

denote the outer and inner control point sets respectively, expressed in the gripper frame. %

Querying the Local Encoder at these locations produces occupancy predictions

\begin{align}
\hat{\mathcal{O}}^{\text{outer}} &= \left\{\hat{o}_i^{\text{outer}}\right\}_{i=1}^{N_o}, \\
\hat{\mathcal{O}}^{\text{inner}} &= \left\{\hat{o}_i^{\text{inner}}\right\}_{i=1}^{N_{\text{in}}},
\end{align}

where $\hat{o}_i^{\text{outer}}, \hat{o}_i^{\text{inner}} \in [0,1]$ are the local occupancy
values $\mathrm{occ}_l$ (Figure~\ref{fig:pipeline}) predicted at
$\mathbf{x}_i^{\text{outer}}$ and $\mathbf{x}_i^{\text{inner}}$, respectively. The refinement objective is defined as

\begin{equation}
\label{eq:refinement_cost}
\mathcal{J}(\xi)
=
w_{\text{free}}
\underbrace{
\frac{1}{N_o}
\sum_{i=1}^{N_o}
\hat{o}_i^{\text{outer}}
}_{\mathcal{L}_{\text{free}}}
+
w_{\text{contact}}
\underbrace{
\left(
1-
\min
\big(
\mathrm{TopK}
(
\hat{\mathcal{O}}^{\text{inner}},
K
)
\big)
\right)
}_{\mathcal{L}_{\text{contact}}}
+
w_{\text{reg}}
\underbrace{
\|\xi\|_2^2
}_{\mathcal{L}_{\text{reg}}},
\end{equation}

where $K < N_{\text{in}}$, and the control points are evaluated at the perturbed grasp pose $G\exp(\xi)$. 

The first term penalizes occupancy at the outer control points and therefore discourages collisions between the gripper and the object. The second term encourages high occupancy at the inner control points, promoting stable contact formation within the finger closing region. Rather than requiring all inner points to lie inside the object, we only require the $K$ most occupied points to exhibit strong occupancy, making the objective more robust to local geometric variations. The final regularization term discourages excessively large pose updates and keeps optimization local to the initial grasp prediction.

Optimization is performed using a sampling-based update procedure. At each iteration, a set of perturbations is sampled from a Gaussian distribution,

\begin{equation}
\xi_i
\sim
\mathcal{N}(\mu,\Sigma),
\end{equation}

where $\mu$ and $\Sigma$ denote the current mean and covariance of the sampling distribution. The sampling distribution is initialized with zero mean and diagonal covariance,

\begin{equation}
\Sigma_0 = \operatorname{diag}\!\big(\sigma_{t,x}^2,\, \sigma_{t,y}^2,\, \sigma_{t,z}^2,\;
\sigma_{r,x}^2,\, \sigma_{r,y}^2,\, \sigma_{r,z}^2\big),
\end{equation}

where $\sigma_{t,\cdot}$ and $\sigma_{r,\cdot}$ are the translational and rotational standard deviations along the axes of the grasp frame. Since perturbations are applied in the grasp frame (Eq.~\ref{eq:exp}), each axis can be scaled independently relative to the gripper, resulting in consistent local refinement across different grasp orientations. In our experiments, we use the values $\sigma_{t,\cdot} = 0.05$ and
$\sigma_{r,\cdot} = 0.1$.

Each sampled perturbation is evaluated using Eq.~(\ref{eq:refinement_cost}), producing a cost value $\mathcal{J}(\xi_i)$ and the resulting costs are converted into importance weights,

\begin{equation}
w_i
=
\frac{
\exp
\left(
-\frac{
\mathcal{J}(\xi_i)-{\mathcal{J}_{\min}}
}{
\lambda
}
\right)
}
{
\sum_j
\exp
\left(
-\frac{
\mathcal{J}(\xi_j)-{\mathcal{J}_{\min}}
}{
\lambda
}
\right)
},
\end{equation}

where $\mathcal{J}_{\min}$ denotes the minimum cost among all sampled perturbations and $\lambda$ controls the sharpness of the weighting distribution. The weighted samples are then used to update the sampling distribution,

\begin{align}
\mu
&=
\sum_i
w_i
\xi_i,
\\
\Sigma
&=
\sum_i
w_i
(\xi_i-\mu)
(\xi_i-\mu)^\top
\end{align}

After each iteration, the covariance is multiplied by a decay factor of $0.9$, gradually concentrating the sampling distribution around promising perturbations and encouraging convergence.

In our implementation, we use $N_{\text{samples}}=12$ perturbations per iteration and perform $N_{\text{iters}}=4$ optimization iterations. After optimization, the final perturbation is applied to the original grasp pose,

\begin{equation}
G_{\text{refined}}
=
G\,\exp(\mu).
\end{equation}

\begin{algorithm}[t]
\caption{Occupancy-Guided Grasp Refinement}
\label{alg:refinement}
\begin{algorithmic}[1]

\Require Initial grasp pose $G$, partial point cloud $\mathcal{P}$, Local Encoder $\Phi$
\Ensure Refined grasp pose $G_{\mathrm{refined}}$

\State Initialize Gaussian perturbation distribution
$(\mu,\Sigma)$

\For{$t = 1,\ldots,N_{\text{iters}}$}

    \State Sample perturbations
    $\{\xi_i\}_{i=1}^{N_{\text{samples}}}
    \sim
    \mathcal{N}(\mu,\Sigma)$

    \For{each perturbation $\xi_i$}

        \State Compute candidate grasp
        $G_i \gets G\exp(\xi_i)$

        \State Extract local crop around $G_i$

        \State Encode crop using Local Encoder $\Phi$

        \State Query occupancies at 
        $\mathcal{O}^{\text{outer}}$
        and
        $\mathcal{O}^{\text{inner}}$

        \State Compute refinement cost
        $\mathcal{J}(\xi_i)$

    \EndFor

    \State Compute importance weights

    \[
    w_i =
    \frac{
    \exp\!\left(
    -\frac{\mathcal{J}(\xi_i)-{\mathcal{J}_{\min}}}{\lambda}
    \right)
    }{
    \sum_j
    \exp\!\left(
    -\frac{\mathcal{J}(\xi_j)-{\mathcal{J}_{\min}}}{\lambda}
    \right)
    }
    \]

    \State Update mean
    $
    \mu \gets \sum_i w_i \xi_i
    $

    \State Update covariance

    \[
    \Sigma \gets
    \sum_i
    w_i
    (\xi_i-\mu)
    (\xi_i-\mu)^\top
    \]

    \State Anneal covariance $ \Sigma \gets 0.9\,\Sigma $

\EndFor

\State $G_{\mathrm{refined}} \gets G\exp(\mu)$

\State \Return $G_{\mathrm{refined}}$

\end{algorithmic}
\end{algorithm}

Algorithm~\ref{alg:refinement} summarizes the complete refinement procedure.

Following refinement, we perform a final occupancy-based validation step to reject grasps that still exhibit collisions or insufficient contact quality. Occupancy values are queried at the same control points used during optimization and a grasp is considered valid only if

\begin{align}
\max(\hat{\mathcal O}_{\text{outer}})
&<
\tau_{\text{free}},
\\
\max(\hat{\mathcal O}_{\text{inner}})
&>
\tau_{\text{contact}},
\end{align}

where $\tau_{\text{free}}=0.2$ and $\tau_{\text{contact}}=0.7$ in all experiments.

This final filtering step removes residual collisions and ensures the existence of sufficiently strong contact evidence within the finger closing region before the grasp is passed to downstream execution.

\subsubsection{Training Procedure}
Training follows the three-stage procedure described in the paper. First, the Global Geometry Encoder and Grasp Generation Module are jointly trained using occupancy, graspability, control-point, and offset supervision. Second, the Dual-Grasp Critic is trained using the force-closure grasp pair annotations provided in DG16M. Finally, the Local Geometry Encoder is trained using local occupancy supervision for occupancy-guided grasp refinement. All networks are optimized using the Adam optimizer. %

\subsection{Inference Pipeline}
Given a segmented RGB-D observation, the object is converted to a partial point cloud and encoded using the Global Geometry Encoder. The Grasp Generation Module predicts dense single-arm grasp candidates together with graspability scores \(p_G\). Candidates with \(p_G > 0.8\) are retained and using farthest-point sampling we obtain $N = 200$ spatially diverse grasp hypotheses. Candidate grasp pairs are then evaluated using the Dual-Grasp Critic, and the highest-scoring grasp pairs are selected. The unique single-arm grasps contained in the top 200 pairs are refined using the Local Geometry Encoder and occupancy-guided \(SE(3)\) optimization to improve contact alignment and reduce collisions, yielding the final \(M=100\) dual-arm grasp predictions.

\begin{figure}[!t]
    \centering
    \includegraphics[width=1\linewidth]{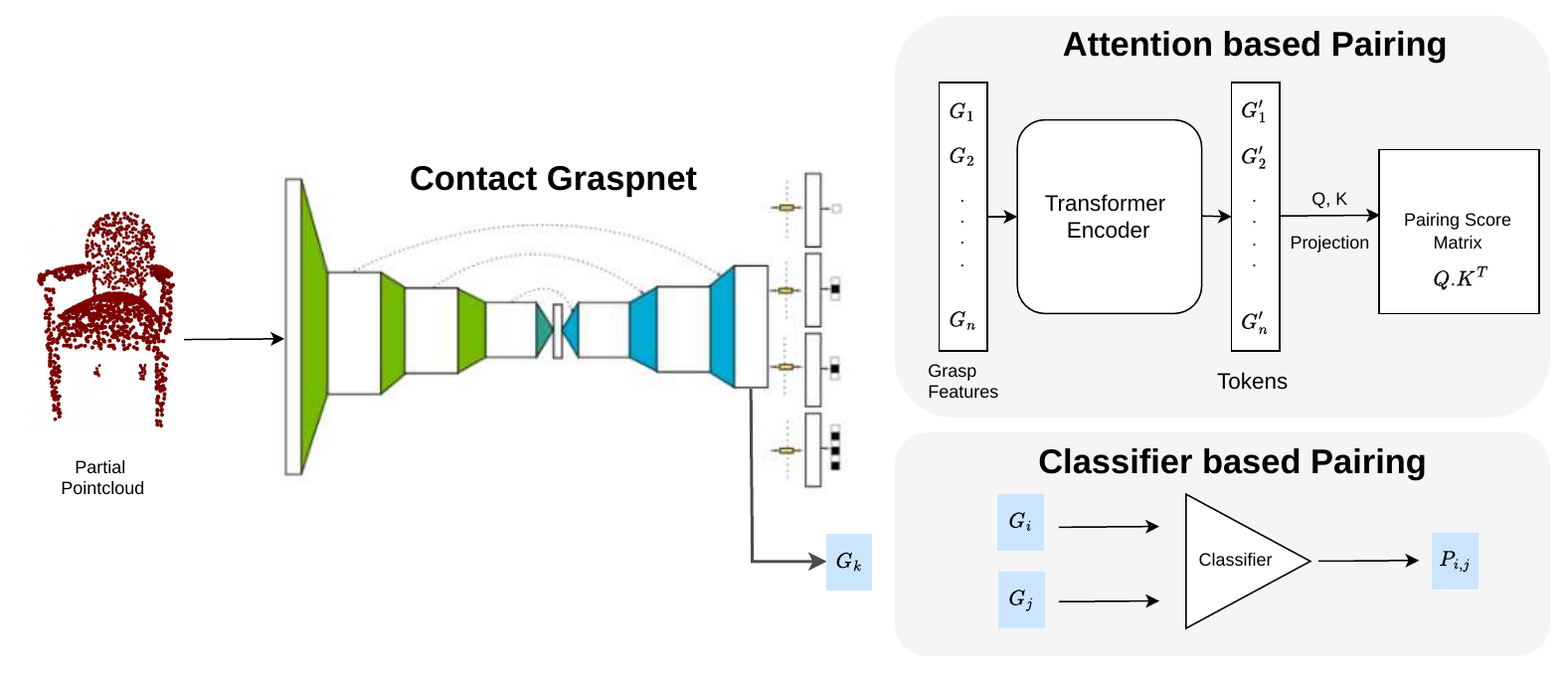}
    \caption{Contact-GraspNet baseline for dual-arm grasp generation. Independent single-arm grasp candidates are generated from a partial point cloud and subsequently paired using either an attention-based transformer architecture or a classifier-based pairing network to identify compatible grasp pairs.}
    \label{fig:cgnet}
\end{figure}

\subsection{Baseline Details} 
\paragraph{ContactGraspNet + Pairing.}
ContactGraspNet (Figure \ref{fig:cgnet}) is first used to generate single-arm grasp candidates from the observed partial point cloud. The high-scoring grasp candidates generated are filtered based on their predicted grasp quality and using farthest-point sampling we obtain \(N=100\) spatially diverse grasp candidates together with their associated point features.

We evaluate two pairing strategies (as shown in Figure \ref{fig:cgnet}):

1. \textit{Attention-Based Pairing.}
The grasp candidates associated point features are processed using a Transformer encoder to obtain contextualized grasp tokens. Query \(Q\) and key \(K\) embeddings   are extracted from the resulting grasp tokens, and pairwise compatibility scores are computed using the interaction \(QK^\top\), yielding an \(N\times N\) grasp-pair score matrix, from which top \(M=100\) pairs are selected. Since ground-truth pair labels are not available for the generated grasp candidates, supervision is generated online using an oracle with access to the object mesh, where each candidate grasp pair is assigned a positive or negative label based on analytical force-closure evaluation.

2. \textit{Classifier-Based Pairing}.
A binary grasp-pair classifier is trained using DG16M force-closure annotations. For each visible observation, ground-truth grasp pairs are matched to the nearest visible ContactGraspNet grasp candidates, and the corresponding point features are concatenated and used to predict whether the pair satisfies force-closure constraints. During inference, the classifier scores all candidate grasp pairs and the top \(M=100\) highest-scoring pairs are retained.

\paragraph{DAGDiff (Partial).}
We train DAGDiff directly on the partial point cloud observations generated from DG16M. Training supervision is restricted to grasp annotations visible in the current observation following the visibility filtering procedure described in Appendix Sec. \ref{sec:Dataset}. During inference, DAGDiff directly predicts dual-arm grasp pairs from the partial point cloud.

\begin{figure}[!t]
    \centering
    \includegraphics[width=1\linewidth]{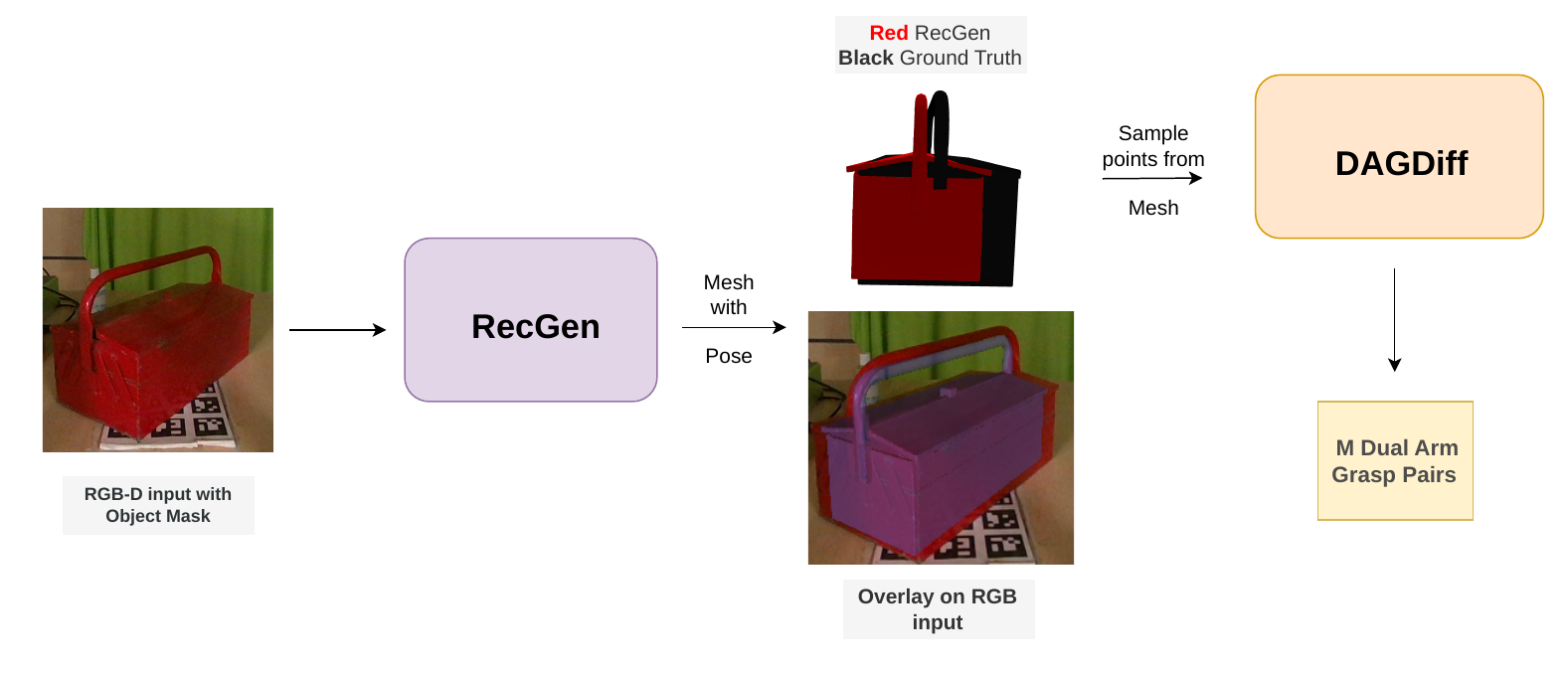}
    \caption{Real-world grasp generation using RecGen and DAGDiff. A complete object mesh is reconstructed from a single RGB-D view using RecGen, and DAGDiff generates dual-arm grasp pairs from points sampled on the reconstructed mesh.}    
    \label{fig:recgen}
\end{figure}

\paragraph{RecGen + DAGDiff(Full).}
We use RecGen to reconstruct a complete object mesh and estimate its pose from a single RGB-D observation. The reconstructed geometry is then provided as input to DAGDiff trained on complete object geometry, which generates the final dual-arm grasp predictions. The pipeline is outlined in Figure \ref{fig:recgen}.

\begin{figure}[!t]
    \centering
    \includegraphics[width=1\linewidth]{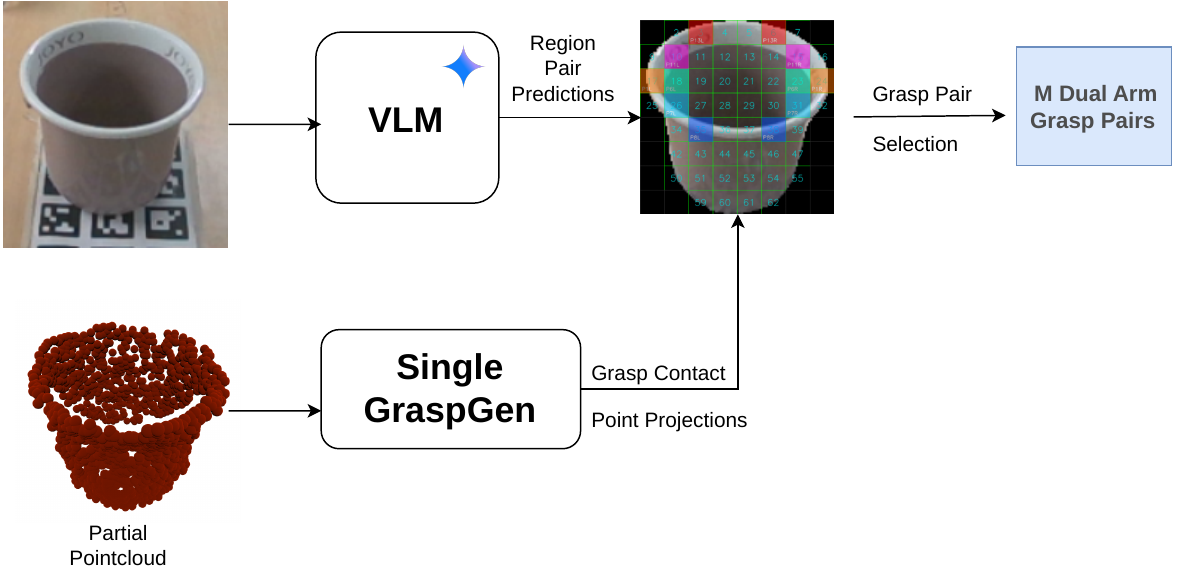}
    \caption{VLM-guided grasp pairing baseline. A vision-language model predicts compatible grasping regions from an image, while a single-arm grasp generator produces grasp candidates from the partial point cloud. Dual-arm grasp pairs are obtained by matching projected grasp contacts to the predicted regions and selecting the top-ranked pairings.}
    \label{fig:vlm}
\end{figure}

\paragraph{Single GraspGen + VLM Pairing.}
We use single-arm grasps generated by our Grasp Generation Module $\mathcal{G}$ and perform grasp pairing using VLM-predicted grasp regions (shown in Figure \ref{fig:vlm}). The VLM predicts pairs of image-space grasp regions represented as bounding-box pairs. Contact locations of the single grasps are projected into the image and assigned to the corresponding regions. Within each region, the top scoring grasp candidates are retained and all pairwise combinations are evaluated using the sum of their grasp scores. The top grasp pairs from each predicted region pair are selected to form the final set of dual-arm grasp predictions. 

\subsection{Results}

\subsubsection{Metrics}
\textit{Force Closure (FC\%), Grasp Success Rate (GS\%), and Pair Collision Rate (PC\%).}
FC\%, GS\%, and PC\% measure the average percentage of predicted grasp pairs that satisfy analytical force-closure constraints, successfully lift and hold the object in simulation, and collide with the object geometry, respectively. Higher FC\% and GS\% indicate better grasp pair quality, while lower PC\% indicates better geometric feasibility.

\textit{Coverage} measures how well the set of force-closure-valid grasp pairs spans the graspable regions of the object.

\begin{equation}
    \mathrm{Coverage}
    =
    \frac{N_{\mathrm{enclosed}}}{N_{\mathrm{graspable}}}
\end{equation}

where $N_{\text{enclosed}}$ is the number of unique positive graspability points lying between the fingers of both grippers, accumulated over all force closure valid grasp pairs, and \(N_{\mathrm{graspable}}\) is the total number of positive graspability points defined in Appendix Sec.~\ref{subsec:Dataset}.

\subsubsection{Comparison with Baselines}
Figure \ref{fig:Qualitative} in the main paper highlights the failure modes of existing approaches under partial observations. Here, we examine these failure cases on a bucket and a chair together with the corresponding predictions from PartialBiGrasp.

\textit{ContactGraspNet + Pairing} generates feasible single-arm grasps but struggles to produce spatially diverse grasps which have accurate contact placement. The resulting grasp pairs are often misaligned with the object geometry, leading to collisions and reduced force-closure performance. These limitations stem from the local point-wise representation, which lacks sufficient geometric context for dual-arm grasp reasoning. In contrast, PartialBiGrasp uses an occupancy-conditioned triplane representation that provides continuous geometric features at arbitrary query locations. This allows grasp generation and pairing to reason over richer geometric context than per-point descriptors, producing more diverse and stable force-closure grasp pairs.

\textit{DAGDiff} trained directly on partial observations is susceptible to occlusion-induced artifacts and often assigns grasps to regions that appear graspable in the visible point cloud but do not correspond to stable grasp contacts. PartialBiGrasp suppresses such false grasp regions (eg. the bucket) through explicit graspability supervision and uses local occupancy reasoning to reject grasps on geometrically infeasible thick structures, as observed on the chair.

The \textit{RecGen + DAGDiff }pipeline is affected by reconstruction and pose estimation errors. Small inaccuracies in the reconstructed geometry propagate to the grasping stage, leading to misaligned contacts around the grasp regions. In contrast, PartialBiGrasp avoids the explicit reconstruction cost and jointly learns geometry and grasp generation, enabling grasp reasoning directly from partial observations.

The \textit{VLM-based pairing} baseline predicts semantically compatible object regions, but semantic compatibility alone does not guarantee a valid dual-arm grasp. We observe that successful grasp pairing requires physical reasoning about how contact locations contribute to stable lifting. Consequently, the selected regions may yield grasp pairs with insufficient separation or contacts on geometrically unsuitable regions.
PartialBiGrasp instead learns physically grounded grasp compatibility from occupancy-conditioned geometric features.

\subsection{Real-World Execution} 
\begin{figure}
    \centering
    \includegraphics[width=0.5\linewidth]{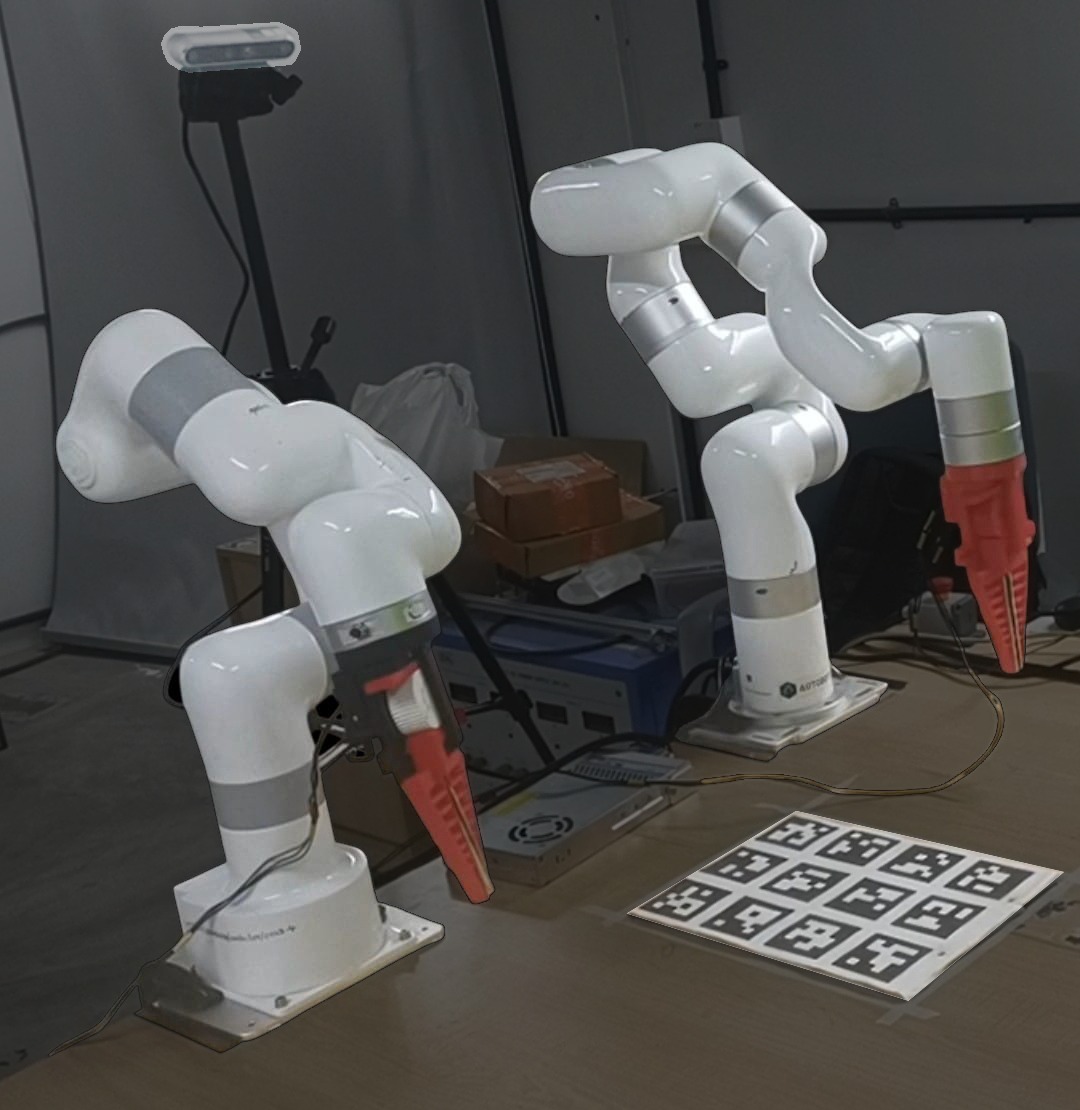}
    \caption{Real-world execution setup consisting of an xArm7 and an xArm6 Lite equipped with parallel-jaw grippers. A RealSense D455 RGB-D camera captures single-view observations of the workspace, and an ArUco board is used for camera extrinsics.}
    \label{fig:real_world_demo}
\end{figure}

All real-world experiments are performed on a heterogeneous dual-arm platform consisting of an xArm7 and an xArm6 Lite equipped with parallel-jaw grippers (Figure \ref{fig:real_world_demo}). A RealSense D455 RGB-D camera provides the single-view RGB-D observations used by the grasping pipeline. Camera calibration is performed using an ArUco marker board placed within the workspace. The generated dual-arm grasps are pruned to retain only collision-free and kinematically reachable grasp pairs before execution on the robot platform. A grasp attempt is considered successful if the object is successfully lifted. The resulting grasps are then executed to evaluate sim-to-real transfer under partial observations and previously unseen object geometries.

\end{document}